\documentclass[journal]{IEEEtran}

\usepackage[colorlinks=true,
            citecolor=blue,
            linkcolor=black,
            urlcolor=blue]{hyperref}

\usepackage{booktabs}
\usepackage{multirow}
\usepackage{graphicx}
\usepackage{amssymb}
\usepackage{makecell}

\usepackage{balance}

\begin{document}
\title{Recent Advances in Multi-modal 3D Intelligence: A Comprehensive Survey and Evaluation}

\author{
Yinjie Lei, Zixuan Wang, Feng Chen, Guoqing Wang, Peng Wang and Yang Yang $^{\dag}$
\IEEEcompsocitemizethanks{
\IEEEcompsocthanksitem Yinjie Lei and Zixuan Wang are with the College of Electronics and Information Engineering, Sichuan University, Chengdu, 610065, China.
E-mail: yinjie@scu.edu.cn, zixuan@stu.scu.edu.cn.
\IEEEcompsocthanksitem Feng Chen is with the School of Computer Science, University of Adelaide, 5005, Adelaide, Australia. 
E-mail: chenfeng1271@gmail.com.
\IEEEcompsocthanksitem Guoqing Wang, Peng Wang and Yang Yang are with the School of Computer Science and Engineering, University of Electronic Science and Technology of China, Chengdu, 611731, China.
E-mail: gqwang0420@hotmail.com, p.wang6@hotmail.com,  yang.yang@uestc.edu.cn.
\IEEEcompsocthanksitem
Corresponding author: Yang Yang (yang.yang@uestc.edu.cn).
}}


\maketitle

\begin{abstract}
Multi-modal 3D Intelligence has gained considerable attention due to its wide applications in autonomous driving and world simulation, etc. Compared to conventional single-modal 3D understanding, introducing an additional modality not only elevates the richness and precision of scene interpretation but also provides a foundation for higher-level physical world interaction. This becomes especially crucial in varied and challenging environments where solely relying on 3D data might be inadequate. While there has been a surge in the development of multi-modal 3D methods over the past six years, especially those integrating multi-camera images (3D+2D) and textual descriptions (3D+language), a comprehensive and in-depth review is notably absent. In this paper, we present a systematic survey of recent progress to bridge this gap. We begin by briefly summarizing the unique challenges among various 3D multi-modal tasks. After that, we present a novel taxonomy that delivers a thorough categorization of existing methods according to modalities and tasks, exploring their respective strengths and limitations. Furthermore, comparative results of recent approaches on several benchmark datasets, together with insightful analysis, are offered.  
Finally, we discuss the unresolved issues and provide several potential avenues for future research.
\end{abstract}


\IEEEpeerreviewmaketitle

\section{Introduction}
\IEEEPARstart{G}{iven} a 3D point cloud and information from another modality, such as 2D images and natural language, multi-modal 3D intelligence seeks to jointly model the geometry, semantics, spatial relationships, and contextual cues of 3D scenes, thereby enabling scene understanding, reasoning, and generation\cite{zeng2023clip2,yang2023zero,yang2023geometry}. A comprehensive modeling of 3D scenes enables agents not only to recognize the categories, locations, and relationships of entities, but also to synthesize novel scene content, layouts, and styles. Compared with using only 3D point clouds, the inclusion of 2D images provides complementary color, texture, and appearance cues, while natural language offers a flexible interface for querying, reasoning about, and generating 3D scenes. Therefore, multi-modal 3D understanding and generation has become a vital research area in computer vision, with broad applications in autonomous driving \cite{mittal2020attngrounder}, robot navigation \cite{wang2019reinforced}, human-computer interaction \cite{bermejo2021exploring}, and world simulation.

Multi-modal 3D intelligence can be broadly categorized according to the auxiliary modality involved. 
\textbf{(1) 3D+2D scene perception.} 
This line of research focuses on integrating 3D geometric measurements, typically from LiDAR point clouds, with 2D appearance cues from camera images to achieve robust scene perception. 
The motivation is that point clouds provide metric depth, object geometry, and spatial layout, while images offer dense texture, color, and semantic context; each modality compensates for the limitations of the other, especially in challenging cases such as distant objects, sparse measurements, occlusion, and adverse lighting conditions \cite{armeni20163d,dai2017scannet,silberman2012indoor,jin2022deformation,geiger2012we,guo2020deep,zhuangwei2021survey,zimmer2022survey,zhou2023zegclip}. 
The central challenge, however, lies in how to align and fuse two heterogeneous representations of the same scene: LiDAR observes sparse 3D points in metric space, whereas cameras capture dense perspective images in the image plane \cite{yeong2021sensor}. 
Existing methods address this challenge through several representative paradigms, including projection-based correspondence, voxel/BEV-level unified representation, attention-based cross-modal interaction, global-to-local fusion, temporal fusion, knowledge distillation, and cross-modal domain adaptation. 
These techniques have substantially advanced 3D object detection and semantic segmentation \cite{xie2020pi,xu2021fusionpainting,zhu2022vpfnet}, making 3D+2D perception a foundational component for autonomous driving, robotic navigation, and large-scale scene understanding.

\begin{figure*}[htbp]
    \centering
    \includegraphics[
        width=\textwidth
    ]{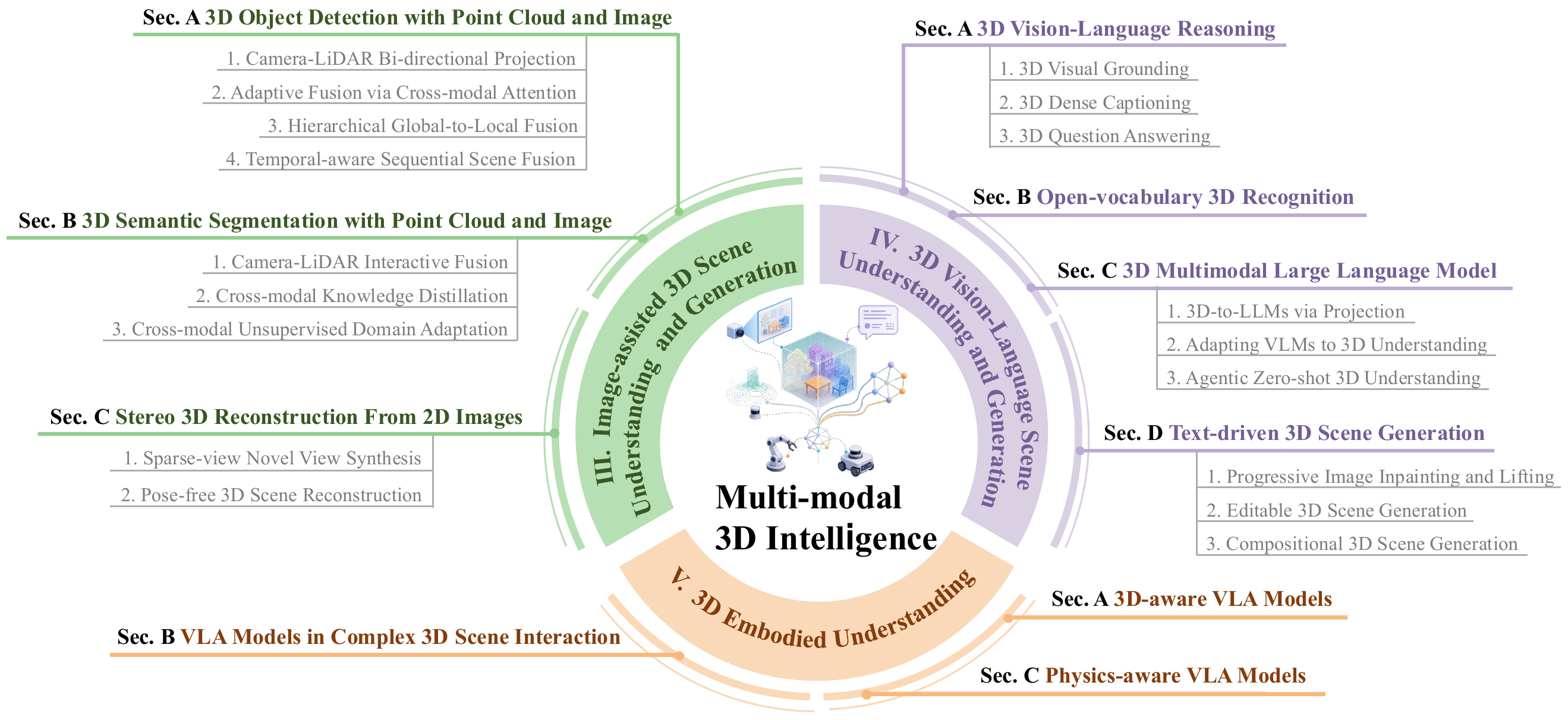}
    \caption{Taxonomy of multi-modal 3D intelligence.}
    \label{fig:overview}
\end{figure*}

\textbf{(2) 3D+language scene understanding and generation.}
Beyond sensor-level fusion, natural language provides a flexible interface for querying, reasoning about, generating, and interacting with 3D scenes. Early 3D+language methods mainly address task-specific vision-language problems, including 3D visual grounding, dense captioning, visual question answering, and open-vocabulary recognition, by aligning linguistic expressions with 3D geometry, object attributes, and spatial relationships through Transformers \cite{wang2023distilling,zhu20233d} or graph neural networks \cite{senior2023graph,jin2021adaptive}. Recently, the rise of LLMs and MLLMs has pushed this field toward general-purpose 3D foundation models, where 3D representations are connected to the semantic and reasoning space of language models for instruction following, open-ended reasoning, and zero-shot generalization. In parallel, language has become a key condition for 3D scene generation, enabling text-driven synthesis, layout editing, and compositional scene creation. More recently, this line of research is further extending toward embodied and action-oriented intelligence, where language commands guide agents to navigate, manipulate, and interact with physical environments through 3D spatial representations. As a result, 3D+language research is evolving from passive scene understanding to a broader paradigm that integrates perception, reasoning, generation, and action.

Despite the fact that numerous methods have emerged in recent years, a large part of the multi-modal 3D intelligence remains rather scattered in different tasks, and no such systematic surveys exist. Therefore, it is necessary to systematically summarize recent studies, comprehensively evaluate the performance of different approaches, and prospectively point out future research directions. This motivates this survey that will fill this gap. The major contributions of this paper can be summarized as: 

\begin{itemize}
    \item \textbf{Systematic survey of multi-modal 3D intellgence.}  To provide readers with a clear comprehension of our article, we categorize algorithms into different taxonomies from the perspective of both required data modality and target downstream tasks, as shown in Figure~\ref{fig:overview}.
    \item \textbf{Comprehensive performance evaluation and analysis.} We compare the existing multi-modal 3D methods on several publicly available datasets. Our in-depth analysis can help researchers in selecting the baseline suitable for their specific applications while also offering valuable insights about the modification of existing methods.
    \item \textbf{Insightful discussion of future prospects.} Based on the systematic survey and comprehensive performance comparison, some promising future research directions are discussed, including large-scale 3D foundation model, persistent 3D memory, computational efficiency of 3D modeling, and scalable 3D representation.
\end{itemize}

\section{Unique Challenges}

Unlike conventional 3D perception, multi-modal 3D intelligence aims to integrate heterogeneous sensory and linguistic signals for scene understanding, generation, and interaction. While additional modalities provide complementary information, they also introduce new challenges that are unique to the 3D multi-modal setting. These challenges arise from the discrepancy between sparse geometry, dense visual appearance, and abstract language semantics, as well as from the need to reason under partial observations and operate in open-ended physical environments. In this section, we summarize several key challenges that commonly appear across existing multi-modal 3D tasks.

\textbf{Heterogeneous Modality Alignment.}
A fundamental challenge in multi-modal 3D intelligence is to align heterogeneous modalities that describe the same physical scene at different levels of abstraction. Point clouds encode sparse metric geometry in 3D coordinates, images provide dense appearance cues in the 2D image plane, and language represents high-level semantic concepts and user intentions through discrete tokens. These modalities differ substantially in coordinate systems, spatial resolution, density, field of view, noise patterns, and semantic granularity. As a result, reliable correspondence must be established not only between 3D points and 2D pixels, but also across views, objects, regions, and linguistic expressions. This alignment is further complicated by calibration errors, occlusions, limited sensor overlap, viewpoint changes, temporal asynchrony, and incomplete observations. Naive fusion strategies, such as direct feature concatenation or one-to-one projection, often fail to capture such complex cross-modal relationships and may introduce noisy or misleading information. 

\textbf{Sparse, Incomplete, and Viewpoint-dependent Observations.}
Another unique challenge of multi-modal 3D intelligence lies in the sparsity, incompleteness, and viewpoint dependence of real-world observations. Unlike 2D images that provide dense visual signals, 3D point clouds are often sparse, irregular, and unevenly distributed, especially for distant, small, or partially occluded objects. Similarly, multi-view images and videos only capture the visible surfaces of a scene, leaving large unobserved regions under-constrained. This partial observability becomes more severe when the input views are sparse, camera poses are unavailable, or the scene contains occlusions and dynamic objects. As a result, models must not only reconstruct or recognize what is directly observed, but also infer missing geometry, maintain object identity across viewpoints, and reason about spatial relationships under uncertainty. A key difficulty is to distinguish reliable observations from plausible but hallucinated completions, which is particularly important for safety-critical applications such as autonomous driving, robotics, and embodied interaction. 

\begin{figure*}[htbp]
    \centering
    \includegraphics[
        width=\textwidth
    ]{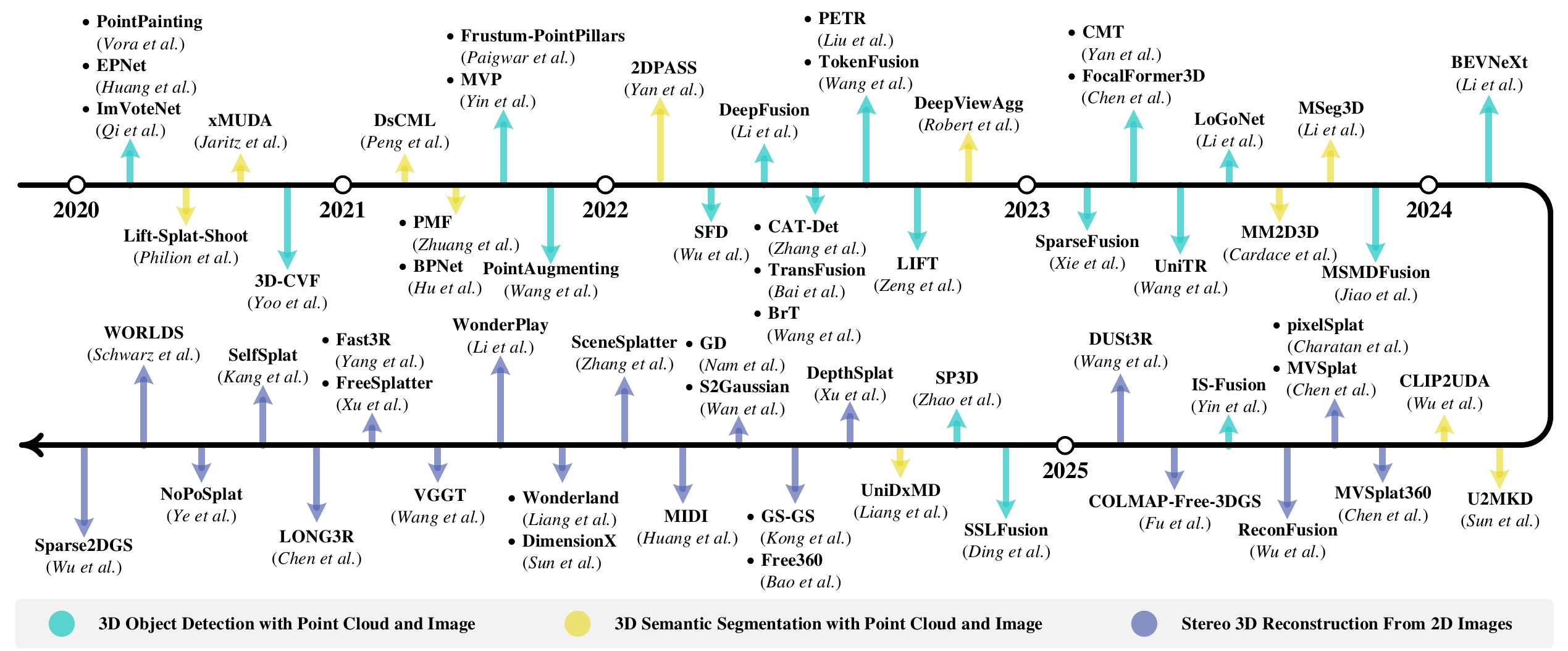}
    \caption{Chronological overview on image-assisted 3D scene understanding and generation approaches from 2020 to the present.
    }
    \label{fig:timeline-1}
\end{figure*}

\textbf{Open-world Generalization.}
Open-world generalization is a critical challenge for multi-modal 3D intelligence, since real-world environments contain unseen categories, diverse scene layouts, long-tail objects, and unconstrained user instructions. Many existing 3D models are still trained and evaluated under closed-set assumptions, relying on predefined object categories, task-specific annotations, or limited scene distributions. Although 2D foundation models, vision-language models, and large language models provide strong semantic priors for open-vocabulary recognition and zero-shot reasoning, transferring these capabilities to 3D scenes remains non-trivial. The difficulty arises from the scarcity of large-scale paired 3D-language data, the domain gap between 2D pretraining and 3D physical environments, and the ambiguity of mapping high-level linguistic concepts to incomplete geometric observations. Moreover, open-world 3D systems must not only recognize novel objects, but also understand unseen spatial relations, object functions, and user intents. 


\textbf{From Perception to Embodied Action.}
A further challenge is to move multi-modal 3D intelligence from passive scene perception toward embodied action. Existing methods often focus on recognizing objects, describing scenes, or answering spatial questions, but embodied agents must use 3D scene understanding to make decisions and interact with the physical world. This requires models to infer not only object categories and locations, but also affordances, reachability, collision constraints, support relationships, object states, and how the scene may change after an action. Such action-oriented understanding is difficult because it depends on persistent scene memory, temporal consistency, physical grounding, and feedback from interaction rather than static visual observations alone. Moreover, language instructions in embodied settings are often ambiguous and must be grounded in both the current 3D scene and the agent's capabilities. 

\section{Image-assisted 3D Scene Understanding  and Generation}
Studies of 3D image-assisted 3D scene understanding and generation can be divided into 3D object detection with point cloud and image, 3D semantic segmentation with point cloud and image, and stereo 3D reconstruction from 2D images. The chronological overview of current 3D image-assisted 3D scene understanding and generation approaches from 2020 to the present is shown in Figure~\ref{fig:timeline-1}.

\subsection{3D Object Detection with Point Cloud and Image}
\begin{figure*}[htbp]
    \centering
    \includegraphics[
        width=\textwidth
    ]{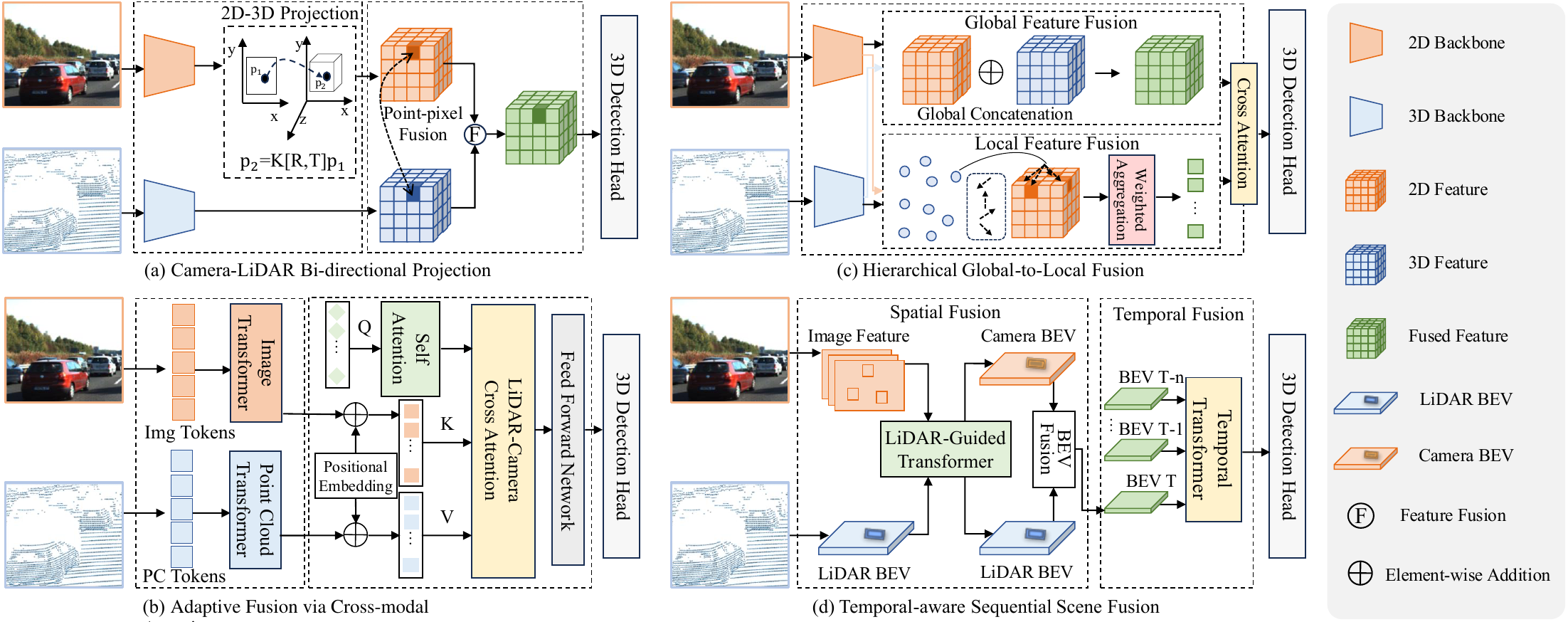}
    \caption{General pipelines of image-assisted 3D object detection. 
    }
    \label{fig:object_detection}
\end{figure*}

\textbf{Camera-LiDAR Bi-directional Projection.} Aligning camera images with LiDAR point clouds requires geometric consistency because LiDAR sensors and RGB cameras operate in diverse coordinate systems. To achieve this, a series of studies utilize camera-LiDAR bi-directional projection to establish a mapping between 3D points and 2D pixels, leveraging camera’s intrinsic parameters (\textit{e.g.,} focal length and principal point coordinates) and extrinsic parameters (\textit{e.g.,} camera position and camera orientation), as shown in Figure~\ref{fig:object_detection} (a). By enforcing such geometric consistency, these techniques enable a one-to-many correspondence between point and pixel features, allowing them to be effectively fused for downstream applications. Early works, such as PointPainting \cite{vora2020pointpainting}, EPNet \cite{huang2020epnet}, EPNet++ \cite{liu2022epnet++}, ImVoteNet \cite{qi2020imvotenet}, MSMDFusion \cite{jiao2023msmdfusion}, and SSLFusion \cite{ding2025sslfusion}, primarily follow a “painting” paradigm. In this paradigm, 2D pixels with their semantic information are lifted into the 3D space to generate virtual points (also known as seeds), and then these virtual points are merged with the original 3D point cloud. These combined features can be used in any LiDAR-only detection approach to produce 3D bounding boxes. Other studies, such as MMF \cite{liang2019multi}, Lift-Splat-Shoot \cite{philion2020lift}, and PointAugmenting \cite{wang2021pointaugmenting} basically utilize a “fetching” paradigm. In this paradigm, 3D LiDAR points are projected onto the 2D image plane via a widely used homogeneous transformation. Subsequently, projected LiDAR points are augmented with the corresponding pixel-wise image features. 

However, such baseline frameworks can only detect large objects near the LiDAR sensor, faraway and small objects are hard to identify because those objects only comprise sparse measurements that even humans cannot clearly see. To detect faraway objects, several methods rely on the 2D image to recognize the category of each faraway object, because the shape of the object does not change sharply with the increase in depth. Then, 2D instance segmentation masks or bounding boxes are extruded into 3D frustums for locating those objects in 3D space. For instance, Frustum-PointPillars \cite{paigwar2021frustum} employs the PointPillars \cite{lang2019pointpillars} to localize smaller objects within the 3D frustum. Faraway-Frustum \cite{zhang2021faraway} predicts the centroid of each 3D frustum via point-cloud clustering to determine potential faraway objects. MVP \cite{yin2021multimodal} further addresses the density disparities between 3D objects located at varying distances by synthesizing dense 3D virtual points around each individual foreground object.

These 3D detection approaches rely on human-annotated labels, which are labor-intensive and prone to annotator bias. As a result, sparsely-supervised 3D object detection has aroused considerable concern, delivering results comparable to fully-supervised detectors with minimal annotation efforts. A general pipeline is deriving semantics from 2D images and mapping semantic labels onto the 3D scene, producing pseudo labels for supervising a 3D detector. However, some challenges continue to exist. On the one hand, the absence of depth information in 2D images may lead to semantic ambiguity around object edges. On the other hand, directly generating pseudo labels often results in incomplete foregrounds, i.e., covering only a fragment of an object instead of the whole. To address such challenges, SP3D \cite{zhao2025sp3d} utilizes a boundary-aware mask shrink operation to obtain accurate seed points for generating bounding box pseudo labels. Besides, SP3D exploits geometric structure of seed points' cross-scale neighborhoods to capture complete coverage of foreground objects. 

\textbf{Adaptive Fusion via Cross-modal Attention.} Typical camera-LiDAR projection approaches often merge features from distinct modalities via simple operations like concatenation or element-wise summation/mean. These approaches regard all features as equally important, failing to account for varying significance of multi-modal features. In real-world scenarios, 2D RGB images usually have noisy information caused by occlusion or overcrowded backgrounds. Directly combining noisy 2D images with 3D point cloud without considering the feature importance may impair 3D detection accuracy. To achieve adaptive cross-modal fusion, many researchers have begun to design novel attention mechanisms in 3D detection, which adaptively fuse features across modalities, selectively emphasizing informative cues while suppressing noisy, as shown in Figure~\ref{fig:object_detection} (b).

Early studies focus on combining 3D point cloud features with 2D image features across the whole scene. BEVFusion \cite{liang2022bevfusion} and PA3DNet \cite{wang2023pa3dnet} perform global fusion via channel attention mechanism (e.g., Squeeze-and-Excitation Network \cite{hu2018squeeze}), which concatenates camera and LiDAR features under the same feature space along channel dimension and recalibrates the importance of each channel in a learnable manner. On this basis, SparseFusion \cite{xie2023sparsefusion} and SFD \cite{wu2022sparse} further crop foreground instances from a whole scene and then separately apply the attention mechanism in each pair of instances, avoiding negative effects from irrelevant objects and background information. However, global channel attention discards spatial information, ignoring differences in pixels and points at different positions. Other approaches, like DeepFusion \cite{li2022deepfusion}, employ the cross-attention mechanism to overcome such issue. Specifically, in the cross-attention mechanism, LiDAR features are regarded as queries, and camera features are served as keys and values. For each query, an attention affinity matrix is computed via inner products between the query and all keys, representing correlations between the current 3D point and all image pixels. Then, values containing camera information are weighted and aggregated using the attention affinity matrix. Compared to channel attention, the cross-attention mechanism can consider similarities of all pixel-point pairs, enabling each 3D point dynamically attend relevant 2D pixels instead of simple one-to-one matching. 

Transformer has emerged as a powerful architecture for processing information from diverse modalities \cite{shen2022aaformer,shvetsova2022everything,zhu2022uni}, due to its inherent ability to align and fuse cross-modal features effectively. Building on such a strength, many endeavors, such as CAT-Det \cite{zhang2022cat}, TransFusion \cite{bai2022transfusion}, BrT \cite{wang2022bridged}, CMT \cite{Junjie2323cross}, and \cite{chen2023focalformer3d}, are dedicated to designing the 3D detection framework based on Transformer architectures. These approaches mainly employ the modal-specific Transformer to encode the intra-modal contextual information, while query-based Transformer performs inter-modal element-wise interactions. Learnable object queries can align the localization of each object across different modalities and encode the object-specific information (\textit{e.g.} size and orientation of box) via separately interacting with the image and point cloud features. For example, CMT \cite{Junjie2323cross} imbues object queries and multi-modal tokens with location awareness. This approach implicitly injects the 3D position information into multi-modal tokens for generating position-aware features and initialize position-guided queries based on the PETR pipeline \cite{liu2022petr}. To address the challenge of false negatives in 3D detection, often stemming from occlusion and cluttered backgrounds, FocalFormer3D \cite{chen2023focalformer3d} introduces the multi-stage hard instance probing technique into the Transformer architecture, which maintains positive regions from previous stages to omit the easy candidates and focus on the hard candidates (false negatives) during training.

Transformer 3D detection architectures described above depend on modality-specific encoders followed by additional query-based fusion, incurring non-negligible computational overheads. Hence, building an efficient Transformer 3D detection model based on the unified multi-modal representation is necessary. UVTR \cite{li2022unifying} performs 3D detection based on the unified voxel representation. First, UVTR maps the image and point cloud into the modal-specific voxel space and formulates the unified voxel representation by adding them together. Then, UVTR samples representative features from the unified voxel space via reference positions and performs instance-level interactions. TokenFusion \cite{wang2022multimodal} effectively combines multiple single-modal Transformers by dynamically detecting useless tokens and substituting these tokens with projected alignment features from the other modality. By selectively substituting tokens, TokenFusion allows the Transformer to learn correlations among multi-modal features efficiently, minimizing redundant computations and maintaining the architecture of the single-modal Transformer unchanged. UniTR \cite{wang2023unitr} processes different modalities through a unified Transformer with shared parameters. Specifically, after transforming the 2D image and 3D point cloud into token sequences, UniTR performs intra-modal representation by a unified Transformer architecture, whose parameters are shared for different modalities. Then, based on 3D-to-2D and 2D-to-3D mapping, UniTR conducts cross-modal representation learning via a modal-agnostic Transformer architecture, simultaneously considering semantic-aware 2D context and geometry-abundant 3D structural relations. 

\textbf{Hierarchical Global-to-Local Fusion.} To simultaneously consider both global context (\textit{e.g.,} relations between each object and a whole scene, as well as interactions among distant objects) and local context (\textit{e.g.,} relations between each object and its surrounding environments) from different modalities, many efforts, like 3D-CVF \cite{yoo20203d}, AutoAlign \cite{chen2022autoalign}, AutoAlignV2 \cite{chen2022autoalignv2}, LoGoNet \cite{li2023logonet}, IS-Fusion \cite{yin2024fusion}, harness the benefits of global fusion and local fusion, as shown in Figure~\ref{fig:object_detection} (c). Taking LoGoNet as an example, in the global fusion stage, image features and point cloud features in a common voxel space are fused by the deformable attention. For each 3D proposal, geometrically aligned pixel features are sampled from image space and fused through the cross-attention mechanism. Finally, information between globally and locally fused features interact with each other via the self-attention mechanism.

\textbf{Temporal-aware Sequential Scene Fusion.} The complementary views and motion cues encoded in the adjacent frames can improve the accuracy of identifying moving 3D objects. Hence, successful 3D object detection, especially in outdoor autonomous driving scenarios, hinges on the optimal exploitation of all available data across sensors and time. Recent advances in sequential modeling demonstrate that Transformer is very competent in modeling the information interaction for sequential data or cross-modal data. The massive amount of 3D points as a sequence input, however, is computationally prohibitive for Transformer. To model the mutual interaction relationship of cross-sensor data over time in a computationally efficient manner, LIFT \cite{zeng2022lift} is proposed. Specifically, LiDAR frames and camera images are encoded as sparsely-located BEV grid features to reduce the computational overload, as well as a sensor-time 4D self-attention module is designed to effectively and efficiently capture the mutual correlations. Instead of directly aggregating sequential cross-sensor data by a 4D self-attention module, BEVFusion4D \cite{cai2023bevfusion4d} separately performs spatial-domain and temporal-domain feature interaction using different modules, as shown in Figure~\ref{fig:object_detection} (d). For spatial-domain feature fusion, 2D image features are mapped into BEV features under the guidance of LiDAR spatial prior via the deformable attention module, and then simply concatenated with the LiDAR BEV features. For aggregating spatially-fused BEV features of consecutive frames, the deformable Transformer-based module dynamically learns the association between temporal features across long spans, aligning temporal features of moving objects without explicit motion calibration. BEVNeXt \cite{li2024bevnext} utilizes a conditional random field to enhance the accuracy of depth perception by integrating depth probabilities with color information. Besides, BEVNeXt expands the receptive field of Res2Net to aggregate historical BEV maps. Finally, the enhanced depth map is used to augment object BEV features for 3D detection.

\begin{figure*}[htbp]
    \centering
    \includegraphics[
        width=\textwidth
    ]{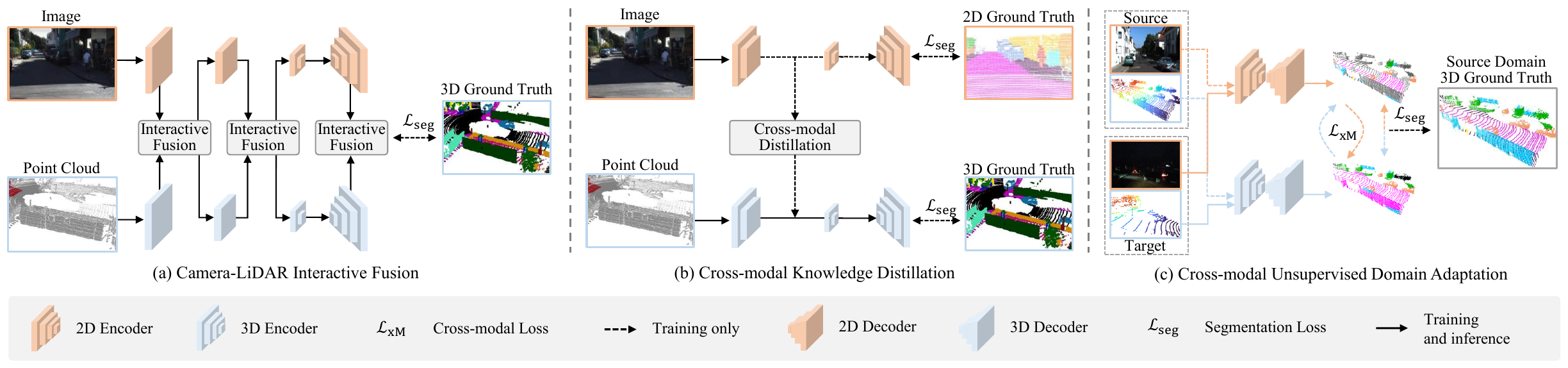}
    \caption{General pipelines of image-assisted 3D semantic segmentation. 
    }
    \label{fig:semantic_segmentation}
\end{figure*}
\subsection{3D Semantic Segmentation with Point Cloud and Image}

\textbf{Camera-LiDAR Interactive Fusion.} On the basis of U-net architecture, a series of studies consider complementarity between 2D image and 3D point cloud data via interactive fusion, and and perform 3D semantic segmentation using fused features, as shown in Figure~\ref{fig:semantic_segmentation} (a). The pioneering framework FuseSeg \cite{krispel2020fuseseg} proposes three key components, forming the foundation of multi-modal 3D segmentation. The first component focuses on separately extracting features from 2D camera images and 3D LiDAR measurements. The second component utilizes the correspondence between LiDAR points and image pixels to concatenate multi-modal features. Finally, fused features are processed by an off-the-shelf segmentation network to generate 3D semantic labels. 

Bridging the differences between 2D images and 3D LiDAR data is important in 3D semantic segmentation, motivating cross-modal fusion aiming to align and integrate complementary features. To mitigate modal discrepancy, PMF \cite{zhuang2021perception} adopts a perception-aware fusion scheme. In such a scheme, 2D image features and 3D point cloud features are mutually projected into each other’s modalities and fused in each modality. Specifically, 3D point clouds are mapped into the camera coordinate system, generating additional spatial-depth information for 2D images. 2D image features are fused to the 3D LiDAR features, injecting the appearance information into the LiDAR measurement. Similarly, in BPNet \cite{hu2021bidirectional}, the projection link matrix between 2D and 3D encoders is built at each level, and the features are transferred bidirectionally according to this link matrix. Lif-Seg \cite{zhao2023lif} is a framework based on the coarse-to-fine fusion. Coarse fusion aims to leverage the contextual information of the 2D image in a one-to-many manner, and the fine-grained fusion can better align the 2D and 3D features by the offset rectification approach.

Effective exploitation of features from multiple views constitutes a core strategy in 3D semantic segmentation, facilitating the integration of complementary information from both modalities. In an effort to effectively aggregate features between multiple views, DeepViewAgg \cite{robert2022learning} introduces a learnable attention-based multi-view aggregation scheme that utilizes the viewing conditions of 3D point cloud to merge the most relevant features from 2D images taken at arbitrary positions. To overcome the limited overlap of the FOV between LiDAR and camera sensors, MSeg3D \cite{li2023mseg3d} employs both geometry-aware and semantic-aware fusion. Geometry-aware fusion module predicts pseudo 2D features to fill in missing camera information and fuses them with LiDAR features in a point-wise geometric correspondence. Then, semantic feature aggregation module combines the LiDAR features and image features into the category-wise semantic embeddings, which describe the typical characteristics of each category from both modalities. Finally, semantic-aware fusion module combines geometry-fused features with semantic embeddings via cross-attention, enabling each point to leverage multi-modal semantic information.

\textbf{Cross-modal Knowledge Distillation.} Interactive fusion approaches require strictly paired 2D image and 3D point cloud data. However, the FOV of LiDAR and cameras only overlap in a small portion, leading to an inability to establish a complete and comprehensive point-to-pixel mapping. For instance, only front-view images are provided by SemanticKITT, while side-view and back-view images are inaccessible \cite{behley2019semantickitti}. Several studies utilize cross-modal knowledge distillation to solve such problem, as shown in Figure~\ref{fig:semantic_segmentation} (b). These methods distill the 2D prior information into the 3D encoder via multi-view distillation based on a small section of overlap multi-modal data during the training stage. During inference, these methods can perform semantic segmentation without 2D visual input, generalizing to non-overlapped data.

The pioneering approach, 2DPASS \cite{yan20222dpass}, employs a fusion-to-single knowledge distillation scheme. Specifically, 2D image features are fused with 3D point cloud features, reserving complete information from multi-modal data. Then, two independent classifiers are separately applied on top of fused and pure 3D features to obtain segmentation scores. Finally, 2DPASS uni-directionally aligns 3D predictions with fusion predictions, transferring the extra 2D information to the 3D encoder as well as maintaining modal-specific information. However, in this scheme, 2D knowledge cannot explicitly be merged into the 3D encoder during training and inference. To address this issue, CMDFusion \cite{cen2023cmdfusion} introduces a 2D knowledge memory branch (implemented by a 3D network), which learns 2D camera features via cross-modality distillation from overlapping points. After learning, such a memory branch can generate pseudo 2D features for an arbitrary 3D point cloud, including points outside the camera FOV, allowing 2D information to be injected without requiring images during inference. In addition, bidirectional fusion block utilizes 2D-to-3D fusion to explicitly enhance the 3D representations by 2D pseudo feature injection, while 3D-to-2D fusion implicitly improves the robustness of 3D features via supervision from the memory branch. With the same motivation, U2MKD \cite{sun2024uni} utilizes bidirectional feature imputation to complete image features for out-of-FOV points and conduct cross-modal fusion. Besides, U2MKD uses a LiDAR-only teacher to guide the cross-modal student network. This design allows a uni-modal teacher to serve as a proxy, delivering additional knowledge such as LiDAR-only data augmentations.

\textbf{Cross-modal Unsupervised Domain Adaptation.} Dense labeling of 3D point clouds is extremely time-consuming. Therefore, unsupervised domain adaptation (UDA), i.e., transferring knowledge learned from source domains with abundant annotations to target domains with unlabeled data, has emerged as a promising research direction for 3D semantic segmentation \cite{wang2018deep, yi2021complete,ding2022doda}. Recent studies mainly consider three real-to-real scenarios: day-to-night (changes in lighting), country-to-country (environmental variations), and dataset-to-dataset (sensor configuration differences). The general pipeline typically involves cross-domain and cross-modal learning, as shown in Figure~\ref{fig:semantic_segmentation} (c). Specifically, cross-domain learning aims to align representations within each modality across domains, either by enforcing consistency between source and target predictions or through feature-level alignment. In contrast, cross-modal learning exchanges information between the 2D and 3D modalities within the same domain, typically by minimizing KL-divergence between their predictions or distilling features from one modality to another.

In multi-modal 3D scenarios, different sensors are affected differently by domain shifts, e.g., cameras are sensitive to lighting changes while LiDAR remains robust. xMUDA \cite{jaritz2020xmuda,jaritz2022cross} addresses this by combining cross-modal learning and self-training. Cross-modal learning lets each modality guide the other, improving robustness for sensitive modalities, while self-training uses high-confidence pseudo-labels on the target domain to adapt to its distribution. Recently, MM2D3D \cite{cardace2023exploiting} incorporates a perception-aware multi-sensor fusion module, injecting depth into the 2D encoder and RGB features into the 3D encoder, which enhances robustness to domain shifts. The effectiveness of this module is because depth information injected into the 2D encoder is less influenced by the domain gap (\textit{e.g.}, lighting conditions change), and RGB information fused into the 3D encoder provides more semantic features (\textit{e.g.}, dark pixels in images acquired at night). 

These methods align multi-modal features in a one-to-one correspondence, which can only achieve information interaction between matched 3D and 2D features, discarding useful information in those unmatched ones. DsCML \cite{peng2021sparse} designs a dynamic sparse-to-dense cross-modal learning to address this issue, where each 3D point interacts with a patch of relevant 2D pixels. The approach leverages the observation that neighboring pixels of the same category are semantically related. Specifically, for each 3D point, DsCML utilizes the deformable convolution \cite{dai2017deformable} to adaptively search the patch of image with appropriate size, dynamically capturing related 2D pixel features with the same category. And, a sparse-to-dense loss is used to align the supremum and infimum of a set of 2D probability scores with their corresponding individual 3D probability score, exchanging the information between different modalities in a many-to-one manner.

However, the above approaches often neglect semantic relationships within heterogeneous data. In essence, cross-modal and cross-domain adaptation seek to capture equivalent information, as semantic concepts remain consistent across modalities and domains, e.g., regardless of changes in appearance or environment, the semantic label of an object remains consistent. Several studies leverage this semantic invariance to address domain shifts. Considering the higher sensitivity of images to environmental changes compared to 3D point clouds, CLIP2UDA \cite{wu2024clip2uda} leverages the frozen CLIP model to facilitate unsupervised domain adaptation. Specifically, given a 2D image from either the source or target domain, the CLIP model produces not only a visual embedding but also a corresponding linguistic embedding representing class semantics, providing domain-invariant semantic guidance. Similarly, UniDxMD \cite{liang2025unidxmd} aims to mitigate coupled domain shifts, such as simultaneous weather variations and LiDAR density changes in autonomous driving. Specifically, UniDxMD quantizes heterogeneous data into a shared discrete space using a weighted combination of local code vectors. Afterwards, UniDxMD aggregates code vectors with the same semantics within this shared discrete space, achieving semantic-aligned knowledge transfer to the target domain.

\subsection{Stereo 3D Reconstruction From 2D Images}
Ongoing advances in neural rendering techniques have significantly improved 3D reconstruction from pose images. The general pipeline is typically modeling a 3D scene with an optimizable neural representation. This differentiable rendering is optimized by minimizing photometric errors between rendered and observed images. Early approaches, such as ReconFusion \cite{wu2024reconfusion} and CAT3D \cite{gao2024cat3d}, represent a 3D scene by Neural Radiance Field (NeRF) \cite{mildenhall2021nerf}, while other approaches use 3D Gaussian Splatting \cite{kerbl3Dgaussians} to model a 3D scene. 

\textbf{Sparse-view Novel View Synthesis.} Novel view synthesis focuses on rendering photorealistic images from viewpoints outside the user-provided views. While NeRF and 3D Gaussian Splatting achieve impressive photometric quality in novel view synthesis given dense observations, they often suffer from significant performance degradation in sparse view scenarios, where optimization becomes severely under-constrained due to insufficient multi-view supervision. To reduce the reliance on dense multi-view captures, recent endeavors have explored feed-forward and prior-guided Gaussian reconstruction from sparse views. A representative line of research is to directly infer 3D Gaussian representations. For example, pixelSplat \cite{charatan2024pixelsplat} is a feed-forward model for directly reconstructing a 3D Gaussian Splatting from sparse images without per scene optimization. Building upon pixelSplat, MVSplat \cite{chen2024mvsplat} designs the cost volume representation for cross-view feature matching, enabling accurate localization of 3D Gaussian centers. DepthSplat \cite{xu2025depthsplat} integrates Gaussian Splatting and depth estimation within a synergistic framework. Specifically, robust depth estimation enhances 3D Gaussian modeling and novel view synthesis, while differentiable Gaussian rendering provides unsupervised signals for learning stronger depth estimator. Another line of work focuses on more challenging wide-baseline and 360$^\circ$ novel view synthesis. For 360$^\circ$ novel view synthesis, MVSplat360 \cite{chen2024mvsplat360} and GS-GS \cite{kong2025generative} leverage the off-the-shelf video/image diffusion model to generate visually realistic unseen and disoccluded views, resolving the challenge posed by minimal overlap among sparse input views. Free360 \cite{bao2025free360} uses layered Gaussian Splatting (near camera foreground layer, far field background layer) to generate unbounded 360$^\circ$ scene, addressing spatial ambiguity inherent under sparse inputs. Beyond improving global geometry and view coverage, some studies aim to enhance reconstruction details under degraded observations. To better capture high-frequency details in feed-forward 3D Gaussian Splatting, GD \cite{nam2025generative} employs the generative densification strategy. This strategy up-samples coarse Gaussian representations and generates their corresponding fine-grained Gaussians in a single forward pass, rather than performing split-and-clone operations on raw Gaussian parameters. S2Gaussian \cite{wan2025s2gaussian} aims to generate a high-quality 3D scene from sparse low-resolution views based on sparse-view high-resolution 3D Gaussian Splatting. Sparse2DGS \cite{wu2025sparse2dgs} focuses on clear 3D surface reconstruction via geometry prioritized enhancement. In particular, Sparse2DGS leverages geometric features derived from multi-view synthesis to initialize Gaussian primitives, and subsequently surpasses appearance overfitting and enforces multi-view geometric consistency during optimization.

Generating 3D scene from a single image is challenging because of inefficient spatial clues captured from an incomplete view. However, the above feed-forward models typically suffer from low-quality reconstructions when inferring geometry from a new image in unseen scenarios, due to the scarcity of supervised information. To address such a challenge, recent methods increasingly exploit various generative priors for single-image 3D scene generation. For example, MIDI \cite{huang2025midi} adapts an off-the-shelf image-conditioned 3D object generation model for compositional scene generation by introducing multi-instance attention mechanism. WORLDS \cite{schwarz2025recipe} regards generation of immersive 3D scene as a context guided 2D inpainting problem. This model uses an image diffusion model to synthesize a coherent panorama, which is subsequently lifted into 3D space through a metric depth estimator. Unobserved regions are further completed by conditioning the inpainting model on rendered point clouds. SceneSplatter \cite{zhang2025scene} generates generic scenes from a single image via the video diffusion model based on momentum paradigm. This approach utilizes rendered images from 3D Gaussian representation as momentum to guide video diffusion model, enhancing video details and maintaining scene consistency. Wonderland \cite{liang2025wonderland} and DimensionX \cite{sun2025dimensionx} equips the video diffusion model with camera conditioning, enabling multi-view consistent scene expansion from a single image by incorporating desired camera tracks. Wonderplay \cite{li2025wonderplay} integrates the physical simulator with video diffusion model for single-image, action-conditioned dynamic 3D scene generation. In particular, Wonderplay utilizes a physical simulator to provide coarse 3D dynamics (flow), which are then used to condition a video diffusion model for fine-grained dynamic synthesis. The generated video facilitates physically plausible 3D scene generation.

\textbf{Pose-free 3D Scene Reconstruction.} Recent pose-free 3D reconstruction studies aim to recover dense scene geometry from uncalibrated images without relying on known camera poses. Early works, such as DUSt3R \cite{wang2024dust3r}, MUSt3R \cite{cabon2025must3r}, and MASt3R \cite{leroy2024grounding} reformulate 3D geometry reconstruction as pointmap regression. Each pointmap densely builds the correspondence between 2D pixels and 3D points. Sparfels \cite{jena2025sparfels} improves computational efficiency through splatted color variance along rays. However, such approaches can only process two images in a single forward pass, and generating 3D scenes from more views requires post-processing to fuse pairwise reconstruction results. In view of this, VGGT \cite{wang2025vggt} can predict camera parameters, depth maps, point maps, and 3D point tracks from up to hundreds of input views in less than a second using large-scale feed-forward Transformer. Similarly, Fast3R \cite{yang2025fast3r} and FreeSplatter \cite{xu2025freesplatter} leverage Transformer architecture to process multiple images in parallel. They utilize global self-attention to fuse visual information across views, reducing error accumulation. Building upon such a feed-forward Transformer architecture, LONG3R \cite{chen2025long3r} introduces a memory mechanism for streaming multi-view 3D scene reconstruction over longer sequences. For each incoming observation, LONG3R selects relevant information from its historical memory, integrates it with the current view features for pointmap prediction, and updates the memory bank accordingly.

\begin{figure*}[htbp]
    \centering
    \includegraphics[
        width=\textwidth
    ]{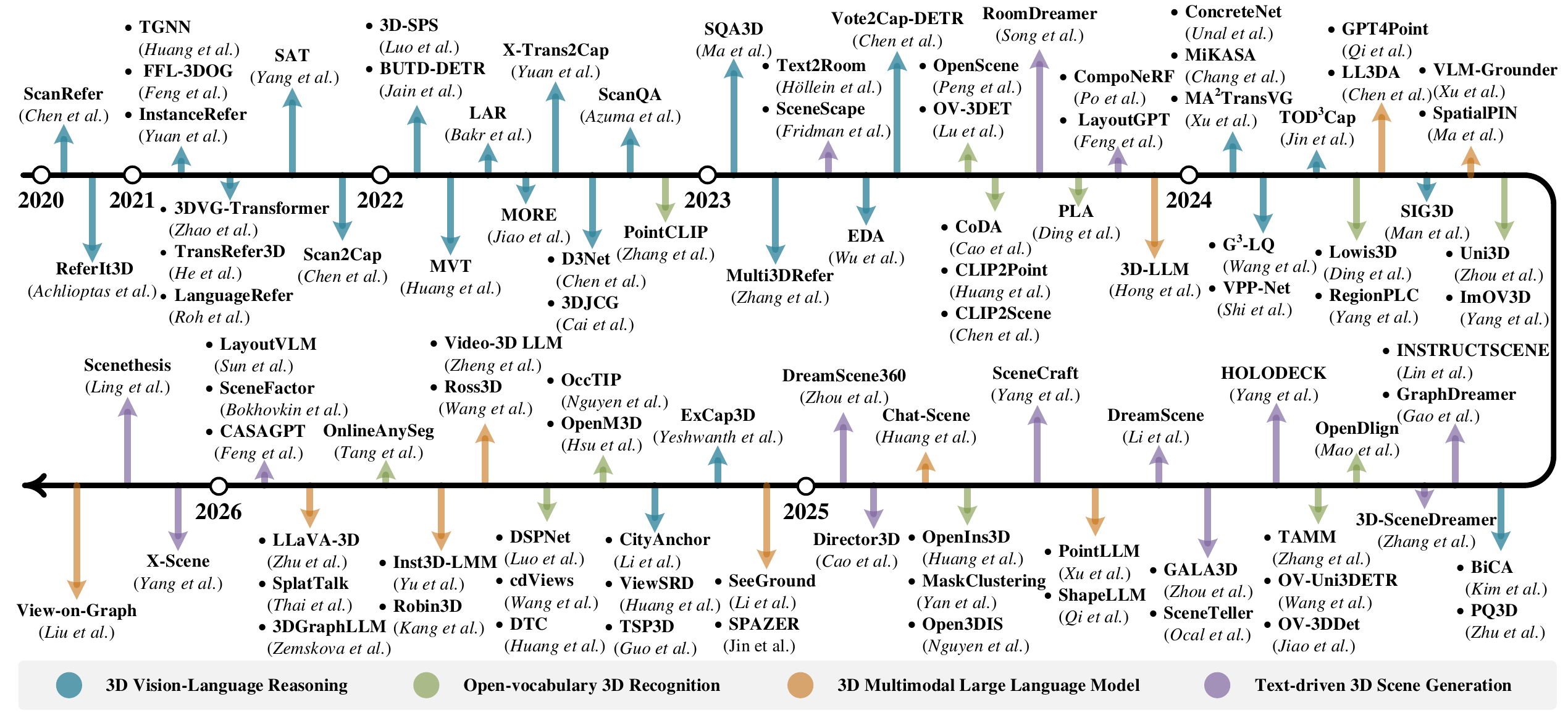}
    \caption{Chronological overview on 3D vision-language scene understanding and generation approaches from 2020 to the present.
    }
    \label{fig:timeline-2}
\end{figure*}

Another line of work explores pose-free reconstruction with 3D Gaussian Splatting. The goal is to recover a renderable 3D representation from unposed images. COLMAP-Free-3DGS \cite{fu2024colmap} simultaneously estimates camera pose and synthesizes novel views from an image sequence, progressively growing the 3D Gaussians set. SelfSplat \cite{kang2025selfsplat} performs pose-free novel view synthesis by integrating 3D Gaussian representation with self-supervised depth and pose estimation. To mitigate errors in pose estimation and inconsistencies in depth prediction, SelfSplat employs a matching-aware pose estimator together with a depth refinement module to improve cross-view geometric consistency. NoPoSplat \cite{ye2025no} anchors one input view’s local camera coordinate system as the canonical space and directly predicts Gaussian primitives for all views within this shared space. This design avoids mapping per-view Gaussians from local coordinates to a global coordinate system, reducing errors caused by inaccurate pose estimation and misalignment among independently predicted Gaussians.

\section{3D Vision-Language Scene Understanding and Generation}
Studies of 3D vision-language scene understanding and generation can be divided into 3D vision-language reasoning, open-vocabulary 3D recognition, 3D multimodal large language model, and text-driven 3D scene generation. The chronological overview of current 3D vision-language scene understanding and generation approaches from 2020 to the present is shown in Figure~\ref{fig:timeline-2}.

\subsection{3D Vision-Language Reasoning}

\textbf{3D Visual Grounding.} Early 3D VG approaches typically employ a dual-stage “detection-then-matching” scheme. In the detection stage, a pre-trained 3D detector \cite{zhou2018voxelnet,he2020structure,pan20213d} or segmenter \cite{engelmann2017exploring,engelmann2019know,xie2020linking} generates a set of candidate 3D object proposals. In the matching stage, visual proposal’s features are combined with linguistic query to identify the desired object. ScanRefer \cite{chen2020scanrefer} pioneered 3D VG by directly concatenating features of the 3D proposal with the linguistic query; however, it barely performs satisfactorily because the cross-modal features are not sufficiently matched. ReferIt3D \cite{achlioptas2020referit3d} and TGNN \cite{huang2021text} encode the relationship among objects by the graph neural network. Extending such a paradigm, FFL-3DOG \cite{feng2021free} and InstanceRefer \cite{yuan2021instancerefer} simultaneously consider object-object relationships and object-to-background global location. Considering such relationships, distractors of the same semantic class can be eliminated. 

Some follow-up endeavors, such as LanguageRefer \cite{roh2022languagerefer}, 3DVG-Transformer \cite{zhao20213dvg}, TransRefer3D \cite{he2021transrefer3d}, MVT \cite{huang2022multi}, ConcreteNet \cite{unal2024four}, MiKASA \cite{chang2024mikasa}, and MA$^{2}$TransVG \cite{xu2024multi}, harness powerful Transformer architecture to achieve granular cross-modal feature interaction and capture long-range spatial relationship. For example, LanguageRefer \cite{roh2022languagerefer}, a BERT-based 3D VG method, formulates 3D VG as a language modeling problem. Each 3D proposal is represented by its semantic category label, which, together with geometric features, is input into the BERT model. Grounding results are then predicted by matching the language description with the semantic labels of the proposals. TransRefer3D \cite{he2021transrefer3d} introduces an entity-and-relation aware Transformer for fine-grained 3D VG. Entity-aware attention aligns object features with corresponding linguistic entity features (e.g., category, color, size), while relation-aware attention captures visual relations between object pairs and aligns them with linguistic relation features (e.g., comparative, spatial relations). MVT \cite{huang2022multi} designs a multi-view Transformer architecture to learn view-robust multi-modal representations. Given a 3D scene from a specific viewpoint, the scene is first rotated into multiple views at equal angular intervals. For each view, proposal features and language features are fused using a Transformer decoder. Finally, information across views is aggregated to identify the target object, thus reducing dependence on the initial viewpoint. In prediction dependent supervised learning, a Transformer model is encouraged to focus on some main attributes (e.g., door) and neglect auxiliary ones (e.g., brown, wooden). To address such bias, MA$^{2}$TransVG \cite{xu2024multi} introduces a multi-attribute aware Transformer. Specifically, the attribute causal analysis module quantifies the contribution of each attribute to the grounding prediction, thereby adaptively assigning attention to all attributes. Guided by this causal analysis, tokens with low attribute attention are exchanged between modalities to enhance cross-modal alignment.

\begin{figure}[tbp]
\centering
\includegraphics[width=\columnwidth]{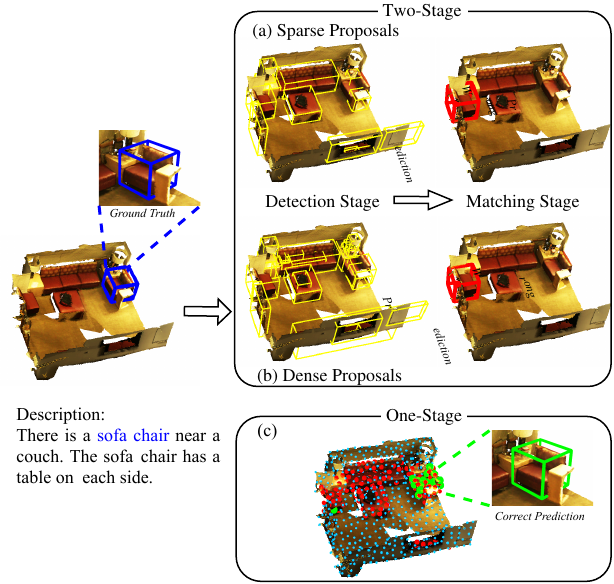}\caption{Diagram of “detection-then-matching” (upper) and “language-guided detection” (bottom) 3D VG approaches \cite{luo20223d}. (a) Sparse proposals may overlook the target in the detection stage. (b) Dense proposals may confuse the matching stage. (c) One-stage approach can progressively select keypoints with the guidance of the language description.}
\label{3D-SPS}
\end{figure}

To mitigate the sparsity and incompleteness inherent in 3D data, SAT \cite{yang2021sat} and LAR \cite{bakr2022look} utilize visual and semantic information from high-quality 2D images to assist 3D VG. Multi3DRefer \cite{zhang2023multi3drefer} is able to identify an arbitrary number (zero, single, multiple) of queried objects within a 3D scene by leveraging a CLIP-like model. CityAnchor \cite{li2025cityanchor} localizes urban objects within city-scale 3D scenes using a VLM-assisted coarse-to-fine matching strategy. UniGround \cite{zhang2026uniground} adapts such a coarse-to-fine matching for learning-free geometric perception, enabling robust generalization to unseen spatial relationships and novel scenes. However, such a dual-stage paradigm faces the dilemma of deciding the proposal numbers. This is because the 3D detector adopted in the detection stage requires sampling a few keypoints to represent the whole 3D scene and generates the corresponding proposal for each keypoint. Sparse proposals may overlook the target in the detention stage, thus the target can not be identified in the matching stage, as shown in Figure~\ref{3D-SPS} (a). Dense proposals may contain unavoidable redundant objects, leading to the inability to distinguish the target in the matching stage because the inter-proposal relationships are too complex, as shown in Figure~\ref{3D-SPS} (b). Furthermore, the keypoint sampling strategy is language-irrelevant, which adds difficulty to the detector for identifying the language-concerned proposal.

To overcome the drawback of “detection-then-matching” scheme, several efforts have proposed one-stage “language-guided detection” framework. This framework unifies object detection and cross-modal feature matching, enabling direct grounding of the desired object, as shown in Figure~\ref{3D-SPS} (c). 3D-SPS \cite{luo20223d} progressively selects keypoints with the guidance of the language description, which makes it easier to locate the query-relevant object. BUTD-DETR \cite{jain2022bottom} pioneers a DETR-like 3D VG model. Follow-up works advance the above one-stage baseline through dense cross-modal alignment, multi-view learning, and inference acceleration. EDA \cite{wu2023eda} and G$^{3}$-LQ \cite{wang2024g} decouple sematic components in a sentence and perform the dense alignment between the fine-grained linguistic cues and 3D point cloud. Since referring sentence frequently convey the relative spatial arrangements, e.g., which are inherently dependent on speaker’s viewpoint. Without viewpoint annotations, leveraging such sentence to localize objects in a 3D environment may lead to ambiguity. In view of this, VPPNet \cite{shi2024aware} explicitly predicts the viewpoint of the speaker, and uses it as an auxiliary input. ViewSRD \cite{huang2025viewsrd} decomposes complex a multi-anchor sentence (encompassing multiple reference objects) into a set of simpler single-anchor statements, and subsequently aggregates multi-view cross-modal features to produce robust 3D grounding. In TSP3D \cite{guo2025text}, a sparse convolutional architecture, along with a language-guide pruning strategy are proposed to achieve efficient 3D VG.

\textbf{3D Dense Captioning.} Unlike 2D image captioning, which primarily generates an overall description of a given image \cite{vinyals2015show,chen2015mind,wu2016value}, 3D dense captioning requires a more nuanced understanding of the location and attribute of objects. Early 3D dense captioning approaches usually follow a dual-stage “detect-then-describe” pipeline, which employs the detection backbone to generate a series of proposals and then create captions for each detected object. Several studies, such as Scan2Cap \cite{chen2021scan2cap}, MORE \cite{jiao2022more}, and SpaCap3D \cite{wang2022spatiality}, TOD${^3}$Cap \cite{jin2024tod3cap} mainly focus on modeling the attributes of each object as well as the relationship between several adjacent objects. For example, Scan2Cap \cite{chen2021scan2cap} regards each proposal as the node in the relational graph to model the relationship between different proposals, generating an enhanced representation of each proposal. Then, Scan2Cap adopts relational and enhanced features to produce more distinctive descriptions through a context-aware attention captioning module. The above methods integrate 2D features (color and texture) into the 3D point cloud via auxiliary encoder. Although this strategy can improve performance, it increases the computational load during the inference stage. To tackle this issue, X-Trans2CaP \cite{yuan2022x} employs a teacher-student knowledge distillation framework. Teacher network leverages 2D-3D inputs, while student uses only 3D inputs. During training, feature consistency constraints guide the student network to learn 2D prior knowledge. As a result, only 3D point cloud needs to be fed into the student network during the inference stage to obtain reliable descriptions that implicitly combine information from both modalities.

Despite the advancements of “detection-then-description” 3D dense captioning pipeline, some problems persist, including the reliance on the detection backbone performance and inadequate consideration of the reciprocal enhancement between detection and captioning. To mitigate the shortcoming of “detection-then-description” scheme, several efforts have proposed one-stage “detection-description parallel”, which regards 3D dense captioning as the set-to-set problem. Vote2Cap-DETR \cite{chen2023end} and Vote2Cap-DETR++ \cite{chen2024vote2cap} represent each 3D object as a unique vote query that encodes its spatial and semantic features, enabling parallel object detection and caption generation within a unified Transformer decoder. However, 3D dense captioning relies on modeling diverse positional relationships across a 3D scene. In a “detection-description parallel” pipeline, precise localization requires local context, while caption generation depends on global context. Consequently, conflicting objectives may arise, as contextual features from multiple spatial positions are integrated into each object representation. To resolve this conflict, BiCA \cite{kim2024bi} introduces a bi-directional contextual attention module, which represents contexts via the weighted aggregation of global objects and enhances object representations by fusing information from non-object contexts. To further improve captioning granularity, ExCap3D \cite{yeshwanth2025excap3d} further extends 3D dense captioning by generating both coarse object-level descriptions and fine-grained statements for individual object parts.

Recently, several studies have demonstrated that 3D VG and dense captioning are naturally complementary. Specifically, 3D VG is relation-oriented, which requires learning relationships among objects for discriminating different objects, especially objects from the same class. On the contrary, 3D dense captioning is object-oriented, which focuses more on learning attributes of the target objects and relationships between the target and its surrounding objects. If 3D VG and dense captioning are integrated into a unified architecture, grounding can provide more relation information to improve captioning performance. Meanwhile, captioning can help enhance grounding performance by providing fine-grained attribute information. Drawing inspiration from module similarity in dual-stage frameworks, some studies, such as D3Net \cite{chen2022d}, 3DJCG \cite{cai20223djcg}, and, PQ3D \cite{zhu2024unifying}, have developed unified architectures for joint 3D VG and dense captioning, which leverage the complementary nature between them to improve performance. For example, 3DJCG introduces a shared-specific framework, which is composed of shared task-agonistic modules and lightweight task-specific modules. First, shared task-agonistic modules learn fine-grained attribute features of each object and complex relations between different objects, which benefit both captioning and VG. Then, lightweight task-specific heads solve dense captioning and VG, respectively, by casting each task as a proxy of another one.

\begin{figure}[t]
\centering
\includegraphics[width=\columnwidth]{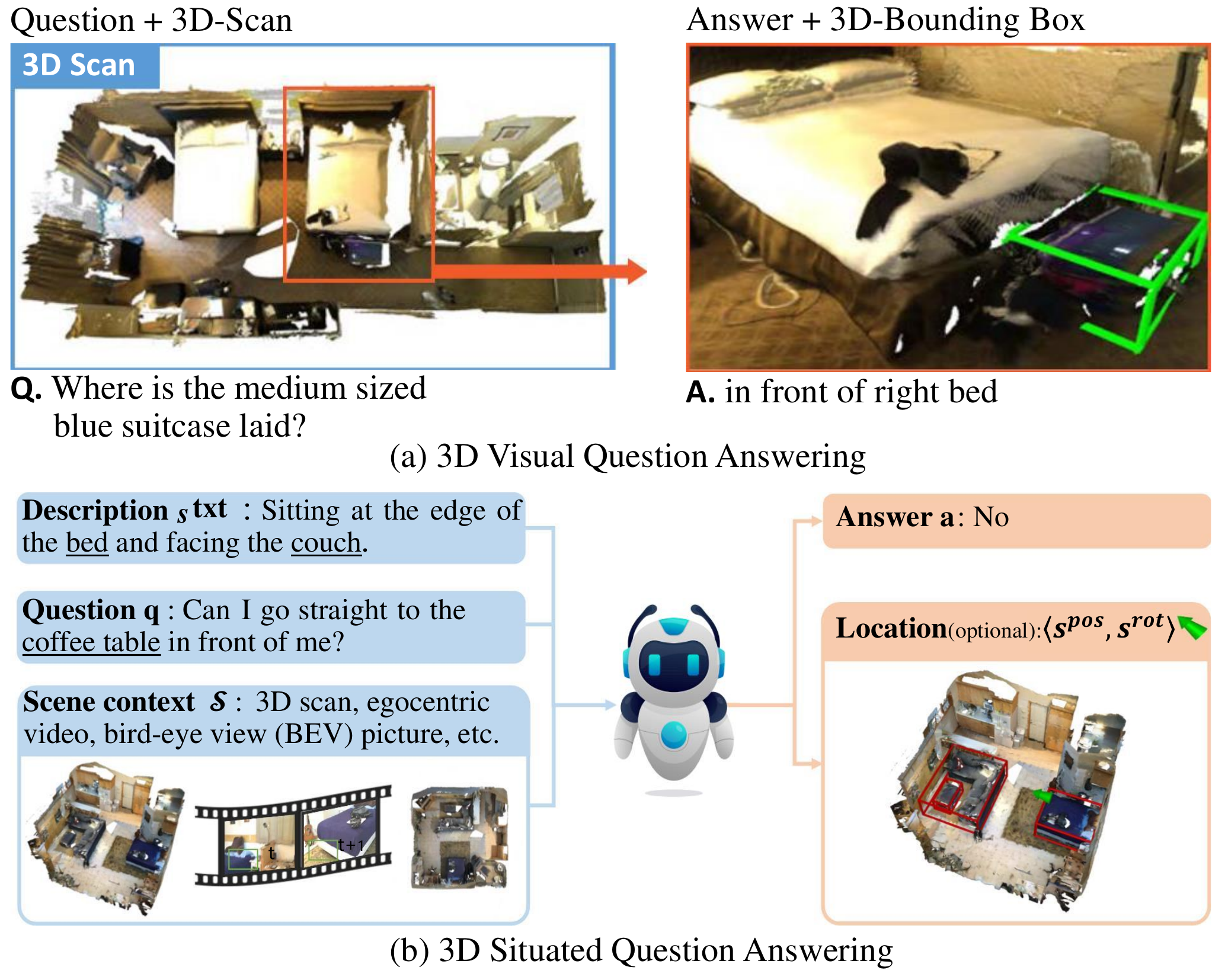}\caption{Illustration of 3D question answering. (a) 3D visual question answering (VQA) \cite{azuma2022scanqa}. This requires predicting an answer phrase and the corresponding 3D bounding boxes according to an entire 3D scan and a linguistic question. (b) 3D situated question answering (SQA) \cite{ma2022sqa3d}. Agents should comprehend and localize the situation of objects in the 3D scene with a given textual description.}
\label{question_answering}
\end{figure}

\textbf{3D Question Answering.} 3D visual question answering (VQA) aims to answer complex questions about rich 3D scenes, which requires models to prossess strong spatial perception and reasoning capabilities. ScanQA \cite{azuma2022scanqa} is the well-known approach designed for 3D VQA, which answers a given question and predicts bounding boxes for described objects in the textual question, facilitating spatial understanding of object direction and location by auxiliary detection loss, as shown in Figure~\ref{question_answering} (a). Extending such as idea, 3D situated question answering (SQA) \cite{ma2022sqa3d} provides the situation description that elucidates the position and orientation of the agent. This requires 3D QA models to understand the given situation from the first-person viewpoint and answer the question under that situation, as shown in Figure~\ref{question_answering} (b). SIG3D \cite{man2024situational} advances this line of research by predicting ego-location and orientation of an agent based on language-aware situation estimator. Simultaneously, 3DGraphQA \cite{wu20243d} models semantic relationships among objects via both question graph and situation graph. To further enhance 3D QA performance, some recent methods leverage multi-view and multi-frame visual representations. DSPNet \cite{luo2025dspnet} integrates multi-view images with 3D point clouds based on language-guided multi-view fusion and adaptive dual-vision perception modules, enabling the perception of both global scene and fine-grained textures. cdViews \cite{wang20253d} automatically selects critical and diverse views, which are subsequently used by multi-frame VLM (LLAVA-OV \cite{li2024llava}) to perform 3D QA. DTC \cite{huang2025zero} reduces over 90\% of visual token usage for multi-frame VLMs by utilizing voxel-based token compression. Driven by the same motivation, SeGPruner \cite{li2026segpruner} utilizes a saliency-aware token selector to preserve semantically crucial visual tokens, a geometry-aware token diversifier promotes wide spatial coverage even when tokens are aggressively reduced.

\subsection{Open-vocabulary 3D Recognition}
In real-world scenarios, a large number of newly acquired 3D data encompasses objects of unseen categories to the deployed 3D recognition models (detection or segmentation models). Even well-performing recognizers may have difficulty in identifying new objects, and re-optimizing models each time an unseen object is encountered is impractical. In 2D vision, similar issues have been alleviated by CLIP’s powerful open-vocabulary recognition capability. A question inevitably arises: Could CLIP be generalized to 3D vision and achieve recognition of unseen 3D objects. Early approaches, PointCLIP \cite{zhang2022pointclip} and OpenDlign \cite{mao2024opendlign} cast raw point clouds onto image planes to generate multi-view images. CLIP’s visual encoder is used to capture multi-view features. And then, zero-shot recognition result for each view is obtained using CLIP’s language encoder.

Following closely, several efforts, such as OpenScene \cite{peng2023openscene}, OV-3DET \cite{lu2023open}, CoDA \cite{cao2023coda}, Uni3D \cite{zhou2024uni3d}, CLIP-FO3D \cite{zhang2023clip}, CLIP$^{2}$ \cite{zeng2023clip2}, CLIP2Point \cite{huang2023clip2point}, CLIP2Scene \cite{chen2023clip2scene}, ImOV3D \cite{yang2024imov3d}, TAMM \cite{zhang2024tamm}, FMOV-3D \cite{zhang2024fm}, OV-Uni3DETR \cite{wang2024ov}, OpenM3D \cite{hsu2025openm3d}, OccTIP \cite{nguyen2025occlusion}, and MPEC \cite{wang2025masked} build specialized 3D encoders by distilling CLIP prior knowledge, thereby expanding well-aligned image-language feature space to embed 3D concepts in unison. For example, in 3D segmentation, OpenScene \cite{peng2023openscene} establishes the correspondence between 2D images (pixels) and 3D scenes (points) based on the available RGB-D data and employs contrastive learning to align the 2D and 3D embeddings. Since image and language features have been already aligned by CLIP, point cloud features are inherently aligned with the language embedding. Another approach, MPEC \cite{wang2025masked} directly enforces both point-to-object consistency and object-to-language alignment, allowing 3D point clouds to inherit CLIP’s language supervision without leveraging 2D images as a bridge. In 3D detection, OV-3DET \cite{lu2023open} uses a dual-stage framework based on the divide-and-conquer paradigm. First, coarse 2D bounding boxes (detected by an open-vocabulary image detector) are used to supervise the 3D detector, teaching it how to locate 3D objects. Second, 3D detector learns to classify unknow objects by leveraging knowledge of CLIP. Other studies, such as PLA \cite{ding2023pla}, Lowis3D \cite{ding2024lowis3d}, RegionPLC \cite{yang2024regionplc}, and OV3D \cite{jiang2024open} yield high-quality language descriptions of 3D scenes through captioning multi-view images, and then connect the 3D visual encoder and CLIP’s language encoder directly using point-discriminative contrastive learning. Instead of leveraging image as a bridge, UniM-OV3D \cite{he2024unim} generates hierarchical descriptions directly from 3D point clouds, including global-view, eve-view, and sector-view descriptions. This provides fine-grained semantic supervisions. 

However, the above approaches based on knowledge distillation requires massive well-aligned 2D images and 3D point clouds for contrastive learning, which are unavailable in real-world scenarios. Given this, OpenIns3D \cite{huang2024openins3d}, MaskClustering \cite{yan2024maskclustering}, Open3DIS \cite{nguyen2024open3dis}, and Zoo3D \cite{lemeshko2025zoo3d} perform zero-shot 3D instance mask prediction based on the off-the-shelf segmentation model, followed by pixel-wise open-vocabulary semantic queries. For example, OpenIns3D acquires class-agnostic masks in the 3D scenes. At the same time, OpenIns3D generates synthetic panoramic images and uses the 2D open-world detector to obtain semantic masks. Afterward, category names are assigned to 3D masks by searching through the results of 2D segmentation. Contrary to OpenIns3D, MaskClustering merges 2D open-vocabulary masks into 3D instances based on the consensus rate between two different views. OnlineAnySeg \cite{tang2025onlineanyseg} extends the paradigm of 2D mask merging to the online 3D segmentation. Nevertheless, these methods primarily rely on geometric consistency to aggregate instance masks. When observed viewpoint is incomplete, association errors may occur, including over-merging of different objects or breakup of a single object. To tackle such an issue, in addition to the geometric constraints, Group3D \cite{kim2026group3d} introduces semantic grouping into the mask clustering process.

\begin{figure*}[htbp]
    \centering
    \includegraphics[
        width=\textwidth
    ]{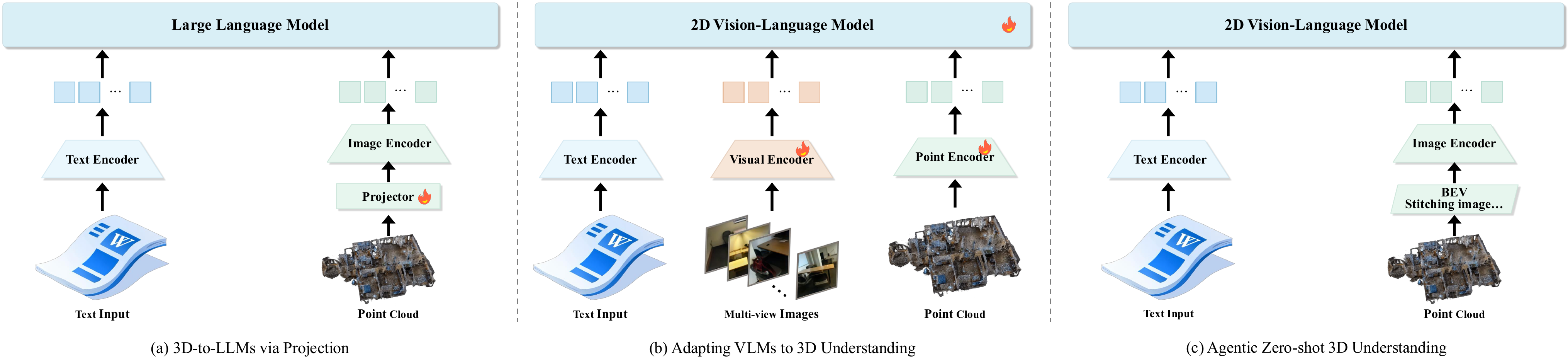}
    \caption{General pipelines of 3D multimodal large language model.
    }
    \label{fig:3d_mllm}
\end{figure*}
\subsection{3D Multimodal Large Language Model}

\textbf{3D-to-LLMs via Projection.} A fundamental challenge in 3D MLLMs lies in the representational discrepancy between 3D features and the embedding space of LLMs. Specifically, 3D features are continuous representations that capture geometric structure, whereas linguistic features are discrete tokens that encode conceptual knowledge. Early approaches, such as PointLLM \cite{guo2023point}, 3D-LLM \cite{hong20233d}, ShapeLLM \cite{qi2024shapellm}, and Kestrel \cite{ahmed2025kestrel} utilize lightweight projection layers to connect the off-the-shelf 3D encoder and LLM, resolving the dimensional and interface mismatch, as shown in Figure~\ref{fig:3d_mllm} (a). Instead of fully connected layer, Mini-GPT-3D \cite{tang2024minigpt}, GPT4Point \cite{qi2024gpt4point}, and LL3DA \cite{chen2024ll3da} use Q-Former as a stronger bridging module, adaptively selecting informative 3D features for LLM reasoning based on learnable queries or visual/linguistic prompts. Whether linear projector or Q-Former, the ability to parse individual objects is lacking. To address such an issue, ChatScene \cite{huang2024chat} and ChatScene++ \cite{huang2026chat} references the 3D scene via object identifiers (linguistic tokens) and represents this scene through object-centric embeddings (2D or 3D tokens). Later, various projectors are proposed to convey complex 3D relations, such as relative position and viewpoint dependency. Chat-3D \cite{wang2023chat} and Robin3D \cite{kang2025robin3d} introduce the relation augmented projector to understand pairwise spatial relationships among objects. S$^2$-MLLM \cite{xu2025s} captures the object correspondences across views through the structure-enhanced module. Inst3D-LLM \cite{yu2025inst3d} gains a more comprehensive spatial understanding via the proposed multi-view cross-modal fusion and 3D instance spatial relation modules. To enable LLMs to perceive egocentric direction (e.g., “an object lies on my right”) and allocentric direction (e.g., “a woman stands on the left of his car”), PoseAlign \cite{liu2026direction} modulates the projection layer of the direction-aware projector with an embedding of 6-DoF pose, while SIG3D \cite{man2024situational} re-encoding 3D features as satiation-guided visual tokens.

However, such architectures rely on standard 3D encoders to capture geometric features projected into the LLM’s input space, resulting in semantic misalignment between geometric and linguistic spaces. This is because 3D encoders are generally pre-trained via self-supervised objectives, such as masked autoencoding and contrastive learning, primarily optimized for geometric discrimination as opposed to semantic understanding. Even when a projection layer is used to bridge the 3D encoder and LLM, simple projectors are often insufficient to achieve a complete semantic conversion. Given such challenges, a “encoder-free” paradigm is emerged, allowing the LLM to case 3D data as a foreign language. LLaVa-3D \cite{zhu2024llava} produces 3D patches by augmenting the 2D CLIP patches with 3D position embeddings. 3DGraphLLM \cite{zemskova20253dgraphllm} drives the LLM via the learnable 3D scene graph representation. SplatTalk \cite{thai2025splattalk} generates 3D tokens from multi-view images suitable for LLM leveraging the 3D Gaussian Splatting. SAGE3D \cite{paul2026point} and ENEL \cite{tang2025exploring} embed point clouds into discrete tokens by a parameter-efficient tokenizer. Scene-LLM \cite{fu2025scene} deploys a plug-and-play scene magnifier module, selecting query-specific 3D object features via leveraging visual preferences of LLM, alleviating interference from redundant tokens in large-scale 3D scene.  

\textbf{Adapting VLMs to 3D Understanding.} Some research has begun to build 3D MLLMs by leveraging the powerful 2D prior (e.g., appearance, lighting properties, and planer geometric cues) from 2D VLMs, as shown in Figure~\ref{fig:3d_mllm} (b). To inject 3D-aware visual supervision into the fine-tuning procedure of the VLMs, Ross3D \cite{wang2025ross3d} uses both cross-view and global-view reconstruction as the pretext objectives. The former aids in understanding relationships between viewpoints through reconstructing masked views by the aggregation of overlapping information from other ones. The later is responsible for modeling the whole 3D scene via recovering BEV images. 

However, 2D VLMs are primarily pre-trained on 2D plane images, which limits their ability to perceive the visual depth required for 3D scene understanding. This has given rise to an intrinsic gap between the visual representations learned by 2D VLMs and the complexity of 3D scene. In addition, the high cost of collecting and annotating large-scale 3D data poses a challenge in fine-tuning 2D VLMs. To solve such a problem, several efforts regard 3D scenes as dynamic videos, enabling the 2D VLMs to directly capture the visual depth from video streams. Video-3D LLM \cite{zheng2025video} pair video frames with the corresponding 3D global coordinates, facilitating the understanding of the relative position between objects and camera. GPT4Scene \cite{qi2025gpt4scene} utilizes rendered BEV images to represent the global scene, and employs STO-markers (object ID) to ensure consistency between the BEV and corresponding video frames. VG-LLM \cite{zheng2026learning} uses VGGT as the 3D visual geometry encoder to improve the robustness against view transformations. Vid-LLM \cite{chen2025vid} recovers real-world geometry from monocular video via the metric depth model. 3D-RFT \cite{linghu20263d} and Scene-R1 \cite{yuan2025scene} apply reinforcement learning to video-driven 3D perception and understanding, optimizing 2D VLMs based on evaluation metrics or predefined rules, such as IoU. Building upon the above efforts, OnlineSI \cite{liu2026onlinesi} further introduces a compact memory representation based on streaming video, making it suitable for long-horizon applications requiring continuous learning from an ever-changing 3D world.

\begin{figure*}[htbp]
    \centering
    \includegraphics[
        width=\textwidth
    ]{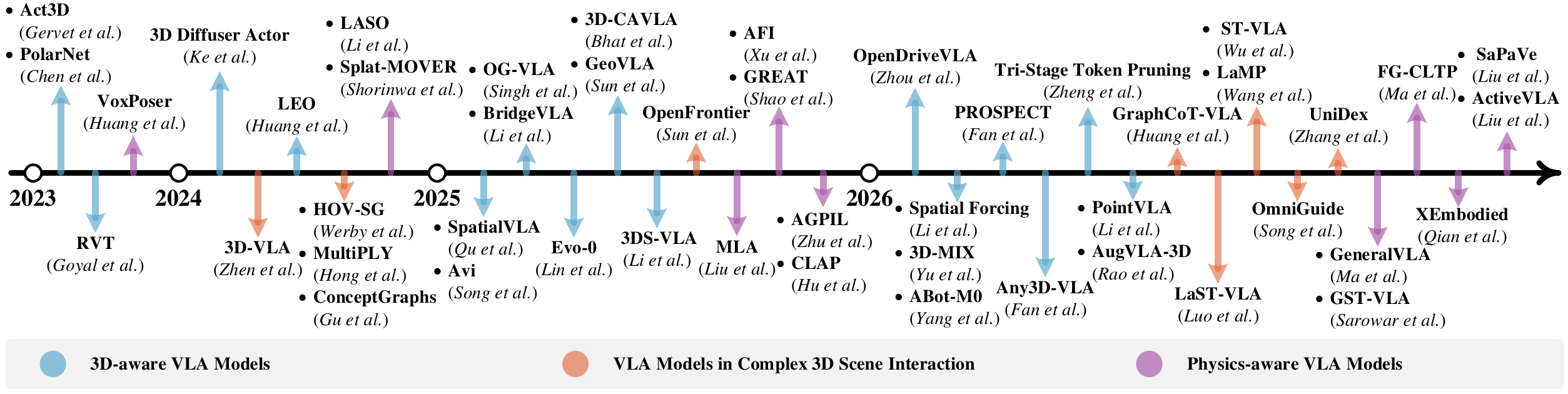}
    \caption{Chronological overview on 3D embodied understanding approaches from 2023 to the present.
    }
    \label{fig:timeline-3}
\end{figure*}

\textbf{Agentic Zero-shot 3D Understanding.} The overwhelming majority of 3D VL understanding approaches depend on supervise learning with annotated 3D datasets and predefined object categories, hindering adaptability and scalability. As a solution, zero-shot approaches utilize 2D VLMs by adapting 3D information into formats compatible VLM processing, primarily including the BEV \cite{zhang2024agent3d}, stitched image sequence \cite{xu2024vlm}, multi-view scene image \cite{lin2025seqvlm}, and query-aligned rendered image \cite{li2025seeground,jin2026spazer}, as shown in Figure~\ref{fig:3d_mllm} (c). However, the above entangled visual representations compel the VLM to process disorganized cues, impairing an effective use of space geometric relationships. To enable the VLM to actively move from less revealing views to more insightful ones, View-on-Graph \cite{liu2026view} organizes 3D scene as a dual-layer scene graph. View and object layers use multi-view images and detected 3D objects as nodes, respectively. Cross-layer edges link views to objects appearing within them. In-layer edges connect geometrically close viewpoints or spatial relationships between objects. Given a query, VLM can perform active exploration along graph connection. Likewise, OpenMap \cite{li2025openmap} aggregates 3D instances from diverse viewpoints to enable zero-shot vision-language navigation through the structural-semantic consensus graph. Of late, multi-agent collaboration and deep thinking have become prominent research directions for zero-shot 3D scene understanding. MAD-3D \cite{zheng2026mag} uses a multi-agent framework for zero-shot 3D grounded reasoning. Planning agent coordinates the reasoning process via dynamically decomposing queries. Grounding agent forms the visual memory by detecting query-specific objects and searching relevant views. Coding agent performs geometric calculation and verification. Such spatial and geometric results are summarized to answer complex questions. SpatialPIN \cite{ma2024spatialpin} performs 3D-aware VQA by progressively prompting with diverse 2D VLMs, such as SAM \cite{kirillov2023segment}, ZoeDepth \cite{bhat2023zoedepth}, and One-2-3-45++ \cite{liu2024one}. PointCoT \cite{zhang2026pointcot} alleviates geometric hallucination by leveraging Chain-of-Thought reasoning based on the Look (perceive 3D geometry), Think (deduce spatial evidence), Answer (deduce the conclusion) paradigm. 

\subsection{Text-driven 3D Scene Generation}

\textbf{Progressive Image Inpainting and Lifting.} Early works, such as SceneScape \cite{fridman2023scenescape} and Text2Room \cite{hollein2023text2room}, generate 3D scenes by incrementally updating a mesh \cite{kato2018neural} (vertices, faces, color of vertices). Given a linguistic description and a camera path, the T2I diffusion model is used to generate novel view images, while a depth estimation model predicts the geometric information of the newly generated images. Based on the view images and their corresponding depths, a unified mesh of the 3D scene is progressively constructed as the camera moves. RoomDreamer \cite{song2023roomdreamer} further repairs low-quality mesh by smoothing holes and correcting distorted geometry. Compared to mesh, NeRF offers a continuous and differentiable 3D scene representation, capturing fine-grained details, while effectively handling regions with depth discontinuities. In Text2NeRF \cite{zhang2024text2nerf} and 3D-SceneDreamer \cite{zhang20243d}, NeRF and T2I diffusion model are combined to generate multi-view consistent 3D scenes via progressive inpainting and updating. However, NeRF requires dense sampling along rays for every pixel, resulting in significantly prolonged rendering times for high-resolution or multi-view images.

Gaussian splatting represents 3D scenes as a collection of parametric Gaussians, each Gaussian encodes the position, size, color, and opacity of a volumetric unit. These Gaussians are projected onto the image plane by direct splatting, mapping each Gaussian as a small disk. Signals of projected Gaussians are subsequently accumulated according to their opacity and radiance to determine the resulting pixel intensities. By avoiding dense per-ray volume sampling, 3D Gaussian splatting approach enables efficient rendering. Building on this, RealmDreamer \cite{shriram2024realmdreamer} and DreamScene \cite{li2024dreamscene,li2025dreamscene} design an end-to-end 3D scene generation framework based on Gaussian splatting. The following approaches build upon the powerful 3D Gaussian splatting architecture, with each proposing novel techniques to overcome distinct challenges. DreamScene360 \cite{zhou2024dreamscene360}, FastScene and SceneDreamer360 \cite{li2024scenedreamer360} utilize panoramic images to synthesize 3D scene, significantly improving global consistency. X-Scene \cite{yang2026x} achieves large-scale 3D driving scene generation by progressively generating new scene components previously generated neighboring regions. Director3D \cite{li2024director3d} aims to address the challenge of complex, unpredictable, and scene-specific camera tracks present in real-world captures. Specifically, Director3D generates dense-view camera tracks from linguistic description by regarding camera parameters as tokens and utilizing a DiT to denoise such tokens conditionally, enabling tracks to accurately follow the description. HunyuanWorld \cite{team2025hunyuanworld} utilizes agentic world layering to automatically decompose complex scenes into semantically meaningful layers, and employs layer-wise 3D modeling to reconstruct each layer independently, enabling the generation of diverse 3D worlds.

\textbf{Editable 3D Scene Generation.} To create user-editable 3D scenes, current approaches focus on modifying parameters of NeRFs, Gaussian Splatting, 3D boxes, or other layout elements within a scene. Set-the-Scene \cite{cohen2023set} and CompoNeRF \cite{bai2023componerf} represent the scene as a composition of object-level NeRFs, each defined with respect to its object proxy. By manipulating the object proxies’ location and shape, the scene can be edited without further fine-tuning. GALA3D \cite{zhou2024gala3d} and SceneTeller \cite{ocal2024sceneteller} introduce 3D Gaussian Splatting for interactive scene editing, addressing the inherent appearance blurring and geometric distortions of NeRF. LayoutGPT \cite{feng2023layoutgpt} instructs large language models to directly generate 3D layouts, providing a foundation for converting the linguistic description into the spatial arrangement. Inspired by LayoutGPT, using “bounding box scene” as a user-friendly layout format, SceneCraft \cite{yang2024scenecraft} creates complex indoor scenes in accordance with user specifications. SceneFactor \cite{bokhovkin2025scenefactor} leverages semantic 3D boxes (category, position, size of each object) to enable controllable 3D scene editing by adding, removing, or resizing these boxes. SDesc3D \cite{feng2026sdesc3d} divide a scene into functional zones, e.g., sleeping and working zones, and use them as implicit spatial anchors to guide layout reasoning. Instead of 3D boxes, CASAGPT \cite{feng2025casagpt} leverages cuboid primitives to represent objects, enabling compact scene synthesis while minimizing object intersections. Then, CASAGPT employs an autoregressive model to sequentially arrange such cuboids to generate physically plausible scenes. In TRELLISWorld \cite{chen2025trellisworld}, global 3D scene synthesis is reformulated as the denoising multiple tiles. Specifically, a scene is decomposed into a set of spatially overlapping regions (also known as tiles). Tiles are independently denoised and subsequently blended by a weighted averaging operation in a single diffusion step. 
LayoutVLM \cite{sun2025layoutvlm} designs differentiable spatial relationships (including, distance, on top of, align with, point towards, against wall), enabling layout optimization via gradient descent.

\begin{figure*}[htbp]
    \centering
    \includegraphics[
        width=\textwidth
    ]{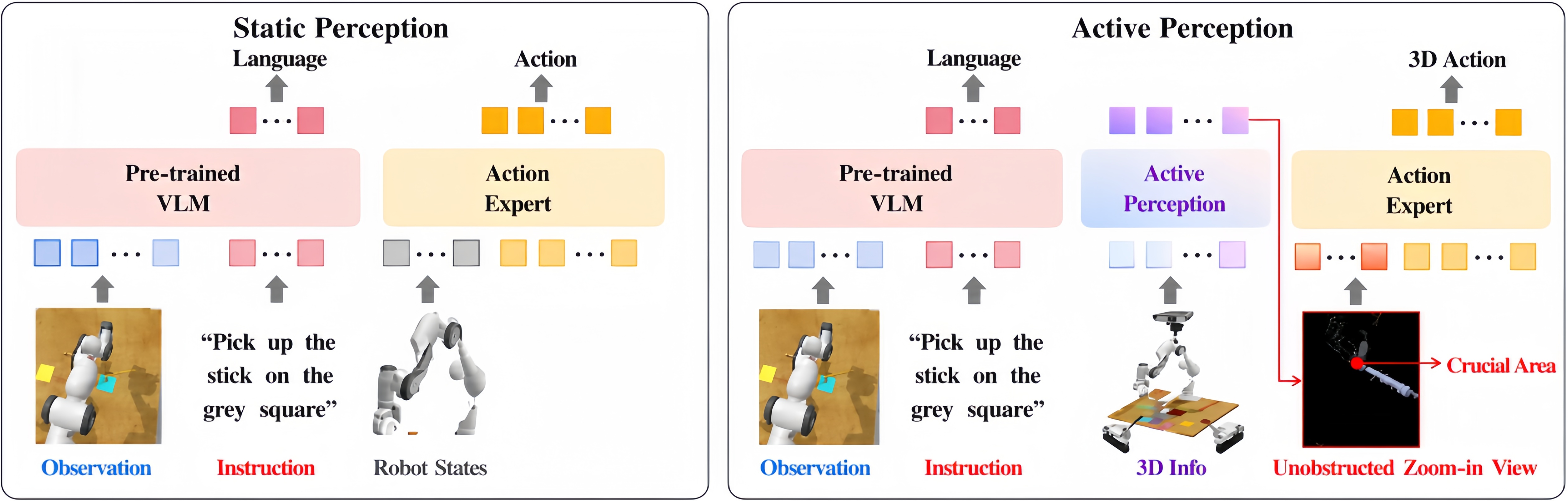}
    \caption{Comparison between static perception and active perception in VLA models. Static perception directly predicts actions from the given observation, instruction, and robot states. Active perception additionally selects informative 3D observations or unobstructed zoom-in views to support more accurate 3D action prediction.}
    \label{fig:vla}
\end{figure*}

\textbf{Compositional 3D Scene Generation.} The main challenge in 3D scene generation approaches lies in their reliance on plain linguistic descriptions to produce a holistic 3D world. When a given description encompasses a complex scene with multiple objects and complicated relationships, simple vectorized linguistic embeddings often fail to accurately capture detailed structure and interactions among objects. In view of this, GraphDreamer \cite{gao2024graphdreamer} and INSTRUCTSCENE \cite{lin2024instructscene} utilize scene graph (derived from linguistic description) to guide the generation of compositional scene. In a scene graph, objects correspond to nodes, and the relationships between any pair of nodes are encoded as edges, enabling decoupled generation of individual objects. Scenethesis \cite{ling2025scenethesis} generates a compositional 3D scene based on a language and vision agentic framework. First, a large language model drafts a coarse scene arrangement. Then, Scenethesis utilizes a vision-language model to generate an image guidance, enabling the searching of appropriate 3D assets and environment maps to assemble a 3D scene. Finally, a physics-aware optimization iteratively adjusts object poses to ensure collision-free and physically plausible placement. HOLODECK \cite{yang2024holodeck} and Reason-3D \cite{berdoz2026text} share a similar framework with Scenethesis. SceneNAT \cite{choi2026scenenat} designs a non-autoregressive DiT framework based on object-wise and attribute-wise masked modeling strategies, enabling the model to learn intra-object and inter-object structures. To enhance relationships among objects, SceneNAT uses a triplet (subject-predicate-object) predictor as the relational guidance.

\section{3D Embodied Understanding}
Studies of 3D embodied understanding can be divided into 3D-aware VLA models, VLA models in complex 3D scene interaction, and physics-aware VLA models. The chronological overview of current 3D embodied understanding approaches from 2023 to the present is shown in Figure~\ref{fig:timeline-3}.

\subsection{3D-aware VLA Models.} VLA models have emerged as a promising avenue for enabling robotic agents to forecast actions from linguistic commands. However, most VLA models rely solely on 2D visual observations, ignoring 3D geometric information that is essential for accurate depth perception and viewpoint-invariant reasoning. Therefore, a key question is how to endow VLA models with spatial understanding of complex 3D physical environments? In view of this, SpatialVLA \cite{qu2025spatialvla} injects the egocentric 3D position encoding into the 2D observation. Act3D \cite{gervet2023act3d}, PolarNet \cite{chen2023polarnet}, 3D-DiffuserActor \cite{ke20243d}, OpenDriveVLA \cite{zhou2026opendrivevla} and Avi \cite{song2025avi} lift multi-view images into 3D space. Conversely, RVT \cite{goyal2023rvt}, OG-VLA \cite{singh2025og} and BridgeVLA \cite{li2026bridgevla} render a 3D scene observation to various canonical orthographic views, lowering computational overhead for action prediction. SpatialForcing \cite{li2025spatial}, Evo-0 \cite{lin2025evo}, 3D-MIX \cite{yu20263d}, ABot-M0 \cite{yang2026abot} implicitly capture 3D geometric features from single or multi-view images based on VGGT encoder. PROSPECT \cite{fan2026prospect} represents 3D spatial features with long-context and absolute scale from streaming videos via CUT3R \cite{wang2025continuous}, yielding impressive navigation performance. ANY3D-VLA \cite{fan2026any3d} utilizes point clouds from simulator, sensor and model-estimated to learn 3D representation which are fused with the corresponding 2D ones. 3D-CAVLA \cite{bhat20253d} and GeoVLA \cite{sun2025geovla} only consider metric 3D point clouds recovered from the depth map. In LEO \cite{huang2023embodied}, the egocentric 2D encoder perceives the agent’s own view, and the object-centric 3D encoder embeds the third-person global perspective. 3DS-VLA \cite{li20253ds} introduces a positional alignment mechanism to fuse 3D point clouds with 2D images. Tri-Stage Token Pruning \cite{zheng20262d} achieves the optimal selection of 2D and 3D tokens. PointVLA \cite{li2026pointvla} and AugVLA-3D \cite{rao2026augvla} directly inject 3D information into the action expert via a lightweight modular block.

\subsection{VLA Models in Complex 3D Scene Interaction.} Beyond developing 3D VLA foundation models, recent advances have focused on enabling robots to perform more sophisticated and long-horizon manipulations in complex environments. Several approaches enhance 3D scene perception. To handle complex scenarios (e.g., multi-floor buildings) or ambiguous commands, ConceptGraphs \cite{gu2024conceptgraphs}, HOV-SG \cite{werby2024hierarchical}, MultiPLY \cite{hong2024multiply}, and GraphCoT-VLA \cite{huang2026graphcot} employ open-vocabulary scene graphs and object-pose graphs, respectively. To enable VLA models to understand complicated 3D scenes, LaST-VLA \cite{luo2026last} uses implicit CoT reasoning. Complementing scene perception, some efforts specialize in performing long-horizon action sequences. 3D-VLA \cite{zhen20243d} envisions goal scenarios to plan actions accordingly with a generative world model. ST-VLA \cite{wu2026st} leverages 4D (3D + time) contexts, enabling online replanning and robust execution of long-horizon robotic manipulation. LaMP \cite{wang2026lamp} leverages 3D scene flow as a motion prior to condition action prediction. Other works emphasize specialized capabilities. OmniGuide \cite{song2026omniguide} improves VLA performance on 3D collision avoidance, physical grounding, articulated object manipulation via unified inference-time guidance. UniDex \cite{zhang2026unidex} focuses on dexterous manipulation. Tailored for 3D navigation, OpenFrontier \cite{sun2025openfrontier} uses visual frontier, i.e., interface between explored and unexplored regions, to serve as physically grounded anchors, balancing exploration and goal-directed behavior.

\subsection{Physics-aware VLA Models.} To improve physical understanding and interaction abilities of robots, a series of works has pursued the following complementary directions. The first direction focuses on implicitly learning physical information. XEmbodied \cite{qian2026xembodied} implicitly learns physical cues from 3D boxes, occupancy grids, and map segmentation. Tex3D \cite{chen2026tex3d} improves the ability of VLA models to withstand physically realizable adversarial attacks, e.g., adversarial 3D textures applied to objects. FG-CLTP \cite{ma2026fg} and MLA \cite{liu2025mla} encode contact states, e.g., force magnitude and principal axis orientation, by leveraging tactile signals as input. The second direction leverages affordances which describe possible interaction opportunities with objects to facilitate action planning. VoxPoser \cite{huang2023voxposer}, LASO \cite{li2024laso}, Splat-MOVER \cite{shorinwa2024splat}, AFI \cite{xu2025affordance}, GREAT \cite{shao2025great}, AGPIL \cite{zhu2025grounding}, GeneralVLA \cite{ma2026generalvla}, and GST-VLA \cite{sarowar2026gst} primarily use the 3D reconstruction technique to generate spatial affordance fields, indicating actionable regions. Based on affordance fields, kinematically valid action sequences can be predicted. Similarly, in CLAP \cite{hu2025generalizable}, the 3D key points serve as the anchors, guiding the VLA models to zoom into salient 3D observations. Further, as shown in Figure~\ref{fig:vla}, SaPaVe \cite{liu2026sapave} and ActiveVLA \cite{liu2026activevla} inject active perception into VLA models, enabling the robot to adaptively select optimal viewpoints during action execution. These viewpoints are characterized by high relevance of amodal object information and minimal occluded regions.

\begin{table*}[t]
\centering
\caption{Performance comparison of 3D object detection approach on KITTI benchmark. The results are reported for Car, Pedestrian, and Cyclist categories under various difficulty levels, including Moderate (Mod), Easy, and Hard. Average Precision (AP) is used as the evaluation metric, and mAP denotes the mean AP over the diverse difficulty levels.}
\label{tab:kitti_detection}
\scriptsize
\setlength{\tabcolsep}{5.2pt}
\renewcommand{\arraystretch}{1.08}
\resizebox{\textwidth}{!}{
\begin{tabular}{ll|rrrr|rrrr|rrrr}
\toprule
\multirow{2}{*}{Method} & \multirow{2}{*}{Venue}
& \multicolumn{4}{c|}{Car}
& \multicolumn{4}{c|}{Pedestrian}
& \multicolumn{4}{c}{Cyclist} \\
\cmidrule(lr){3-6} \cmidrule(lr){7-10} \cmidrule(lr){11-14}
& & Mod & Easy & Hard & mAP
& Mod & Easy & Hard & mAP
& Mod & Easy & Hard & mAP \\
\midrule

\multicolumn{14}{l}{\textit{Camera-LiDAR Bi-directional Projection}} \\



Frustum-PointPillars \cite{paigwar2021frustum} & ICCV2021
& - & - & - & -
& 42.89 & 51.22 & 39.28 & 44.46
& - & - & - & - \\

Faraway-Frustum \cite{zhang2021faraway} & ITSC2021
& 79.05 & 87.45 & 76.14 & 80.88
& 38.58 & 46.33 & 35.71 & 40.21
& 62.00 & 77.36 & 55.40 & 64.92 \\

EPNet++ \cite{liu2022epnet++} & TPAMI2022
& 81.96 & 91.37 & 76.71 & 83.35
& 44.38 & 52.79 & 41.29 & 46.15
& 59.71 & 76.15 & 53.67 & 63.18 \\

SSLFusion \cite{ding2025sslfusion} & AAAI2025
& 81.36 & 90.23 & 76.56 & 82.72
& 44.72 & 52.39 & 42.35 & 46.49
& 62.32 & 78.06 & 56.49 & 65.62 \\

\midrule

\multicolumn{14}{l}{\textit{Adaptive Fusion via Cross-modal Attention}} \\
PA3DNet \cite{wang2023pa3dnet} & TII2023
& 82.57 & 90.49 & 77.88 & 83.65
& - & - & - & -
& - & - & - & - \\

SFD \cite{wu2022sparse} & CVPR2022
& 84.76 & 91.73 & 77.92 & 84.80
& - & - & - & -
& - & - & - & - \\

CAT-Det \cite{zhang2022cat} & CVPR2022
& 81.32 & 89.87 & 76.68 & 82.62
& 45.44 & \textbf{54.26} & 41.94 & 47.21
& 68.81 & 83.68 & 61.45 & 71.31 \\

\midrule

\multicolumn{14}{l}{\textit{Hierarchical Global-to-Local Fusion}} \\

LoGoNet \cite{li2023logonet} & CVPR2023
& \textbf{85.06} & \textbf{91.80} & \textbf{80.74} & \textbf{85.87}
& \textbf{47.43} & 53.07 & \textbf{45.22} & \textbf{48.57}
& \textbf{71.70} & \textbf{84.47} & \textbf{64.67} & \textbf{73.61} \\

\bottomrule
\end{tabular}}
\end{table*}

\begin{table*}[t]
\centering
\caption{Performance comparison of 3D object detection approaches on nuScenes benchmark. ``L'', ``C'', and ``T'' denote LiDAR, camera, and time information, respectively. The results are reported in terms of mAP, NDS, and AP.}
\label{tab:nuscenes_detection}
\tiny
\setlength{\tabcolsep}{2.4pt}
\renewcommand{\arraystretch}{1.08}
\resizebox{\textwidth}{!}{
\begin{tabular}{lll|rr|rrrrrrrrrr}
\toprule
Method & Modality & Venue & mAP & NDS
& Car & Truck & C.V. & Bus & Trailer & Barrier & Motor. & Bike & Ped. & T.C. \\
\midrule

\multicolumn{15}{l}{\textit{Camera-LiDAR Bi-directional Projection}} \\

MVP \cite{yin2021multimodal} & LC & NeurIPS2021 & 66.4 & 70.5
& 86.8 & 58.5 & 26.1 & 67.4 & 57.3 & 74.8 & 70.0 & 49.3 & 89.1 & 85.0 \\

PointAugmenting \cite{wang2021pointaugmenting} & LC & CVPR2021 & 66.8 & 71.0
& 87.5 & 57.3 & 28.0 & 65.2 & 60.7 & 72.6 & 74.3 & 50.9 & 87.9 & 83.6 \\

MSMDFusion \cite{jiao2023msmdfusion} & LC & CVPR2023 & 71.5 & 74.0
& 88.4 & 61.0 & 35.2 & 71.4 & 64.2 & 80.7 & 76.9 & 58.3 & 90.6 & 88.1 \\

\midrule

\multicolumn{15}{l}{\textit{Adaptive Fusion via Cross-modal Attention}} \\
BEVFusion \cite{liang2022bevfusion} & LC & NeurIPS2022 & 71.3 & 73.3
& 88.5 & 63.1 & 38.1 & 72.0 & 64.7 & 78.3 & 75.2 & 56.5 & 90.0 & 86.5 \\

SparseFusion \cite{xie2023sparsefusion} & LC & ICCV2023 & 72.0 & 73.8
& 88.0 & 60.2 & 38.7 & 72.0 & 64.9 & 79.2 & 78.5 & 59.8 & \textbf{90.9} & 87.9 \\

\midrule

\multicolumn{15}{l}{\textit{Cross-modal Transformer-based Methods}} \\
UVTR \cite{li2022unifying} & LC & NeurIPS2022 & 67.1 & 71.1
& 87.5 & 56.0 & 33.8 & 67.5 & 59.5 & 73.0 & 73.4 & 54.8 & 86.3 & 79.6 \\

TransFusion \cite{bai2022transfusion} & LC & CVPR2022 & 68.9 & 71.7
& 87.1 & 60.0 & 33.1 & 68.3 & 60.8 & 78.1 & 73.6 & 52.9 & 88.4 & 86.7 \\

CMT \cite{Junjie2323cross} & LC & ICCV2023 & 72.0 & 74.1
& 88.0 & 63.3 & 37.3 & \textbf{75.4} & 65.4 & 78.2 & 79.1 & \textbf{60.6} & 87.9 & 84.7 \\

FocalFormer3D \cite{chen2023focalformer3d} & LC & ICCV2023 & 71.6 & 73.9
& 88.5 & 61.4 & 35.9 & 71.7 & 66.4 & 79.3 & 80.3 & 57.1 & 89.7 & 85.3 \\

UniTR \cite{wang2023unitr} & LC & ICCV2023 & 70.9 & 74.5
& 87.9 & 60.2 & 39.2 & 72.2 & 65.1 & 76.8 & 75.8 & 52.2 & 89.4 & \textbf{89.7} \\

\midrule

\multicolumn{15}{l}{\textit{Hierarchical Global-to-Local Fusion}} \\
3D-CVF \cite{yoo20203d} & LC & ECCV2020 & 42.2 & 49.8
& 79.7 & 37.9 & - & 55.0 & 36.3 & 47.1 & 37.2 & - & 71.3 & 40.8 \\

AutoAlign \cite{chen2022autoalign} & LC & IJCAI2022 & 65.8 & 70.9
& - & - & - & - & - & - & - & - & - & - \\

AutoAlignV2 \cite{chen2022autoalignv2} & LC & ECCV2022 & 68.4 & 72.4
& 87.0 & 59.0 & 33.1 & 69.3 & 59.3 & - & 72.9 & 52.1 & 87.6 & - \\

IS-Fusion \cite{yin2024fusion} & LC & CVPR2024 & 73.0 & \textbf{75.2}
& 88.3 & 62.7 & 38.4 & 74.9 & \textbf{67.3} & 78.1 & \textbf{82.4} & 59.5 & 89.3 & 89.2 \\

\midrule

\multicolumn{15}{l}{\textit{Temporal-aware Sequential Scene Fusion}} \\
LIFT \cite{zeng2022lift} & LC & CVPR2022 & 65.1 & 70.2
& 87.7 & 55.1 & 29.4 & 62.4 & 59.3 & 69.3 & 70.8 & 47.7 & 86.1 & 83.2 \\

BEVFusion4D \cite{cai2023bevfusion4d} & LCT & arXiv2023 & \textbf{73.3} & 74.7
& \textbf{89.7} & \textbf{65.6} & \textbf{41.1} & 72.9 & 66.0 & \textbf{81.0} & 79.5 & 58.6 & \textbf{90.9} & 87.7 \\

\bottomrule
\end{tabular}}
\end{table*}

\begin{table*}[t]
\centering
\caption{Performance comparison of 3D semantic segmentation approaches on KITTI, nuScenes, Waymo, and ScanNet benckmarks. mIoU and (or FwIoU) are reported.}
\label{tab:semantic_segmentation}
\tiny
\setlength{\tabcolsep}{2.8pt}
\renewcommand{\arraystretch}{1.08}
\resizebox{\textwidth}{!}{
\begin{tabular}{@{}ll|cc|ccc|cc|cc@{}}
\toprule
Method & Venue
& \multicolumn{2}{c|}{SemanticKITTI}
& \multicolumn{3}{c|}{nuScenes}
& \multicolumn{2}{c|}{Waymo}
& \multicolumn{2}{c}{ScanNet} \\
& & Test mIoU & Val mIoU
& Test mIoU & Test FwIoU & Val mIoU
& Test mIoU & Val mIoU
& mIoU & 6-Fold mIoU \\
\midrule

\multicolumn{11}{@{}l}{\textit{Camera-LiDAR Interactive Fusion}} \\

PMF \cite{zhuang2021perception} & ICCV2021
& - & 63.9
& 77.0 & 89.0 & 76.9
& - & 58.2
& - & -  \\

BPNet \cite{hu2021bidirectional} & CVPR2021
& - & -
& - & - & -
& - & -
& \textbf{74.9} & -  \\

LIF-Seg \cite{zhao2023lif} & IEEE TMM2021
& - & -
& - & - & 78.2
& - & -
& - & -  \\

DeepViewAgg \cite{robert2022learning} & CVPR2022
& - & -
& - & - & -
& - & -
& 71.0 & - \\

MSeg3D \cite{li2023mseg3d} & CVPR2023
& - & 66.7
& 81.1 & 91.4 & 80.0
& 70.5 & 69.6
& - & -  \\

\midrule

\multicolumn{11}{@{}l}{\textit{Cross-modal Knowledge Distillation}} \\
2DPASS \cite{yan20222dpass} & ECCV2022
& \textbf{72.9} & 68.6
& 80.8 & 90.1 & 79.4
& - & -
& - & -  \\

CMDFusion \cite{cen2023cmdfusion} & arXiv2023
& 71.6 & 66.2
& 80.8 & 90.3 & 77.3
& - & -
& - & -  \\

U2MKD \cite{sun2024uni} & IEEE TPAMI2024
& - & \textbf{69.6}
& \textbf{84.2} & \textbf{91.7} & \textbf{83.1}
& \textbf{70.6} & \textbf{73.0}
& - & -  \\

\bottomrule
\end{tabular}}
\end{table*}

\begin{table*}[t]
\centering
\caption{Performance comparison of 3D visual grounding approaches on ScanRefer benchmark. Accuracy is evaluated by Acc@0.25IoU and Acc@0.5IoU.}
\label{tab:visual_grounding}
\tiny
\setlength{\tabcolsep}{3.4pt}
\renewcommand{\arraystretch}{1.08}
\resizebox{\textwidth}{!}{
\begin{tabular}{ll|rr|rr|rr}
\toprule
Method & Venue
& \multicolumn{2}{c|}{Unique}
& \multicolumn{2}{c|}{Multiple}
& \multicolumn{2}{c}{Overall} \\
& & Acc@0.25 & Acc@0.5 & Acc@0.25 & Acc@0.5 & Acc@0.25 & Acc@0.5 \\
\midrule

\multicolumn{8}{l}{\textit{3D Visual Grounding Expert Models}} \\
MVT \cite{huang2022multi} & CVPR2022 & 77.67 & 66.45 & 31.92 & 25.26 & 40.80 & 33.26 \\
3DJCG \cite{cai20223djcg} & CVPR2022 & 78.75 & 61.30 & 40.13 & 30.08 & 47.62 & 36.14 \\
3D-SPS \cite{luo20223d} & CVPR2022 & 81.63 & 64.77 & 39.48 & 29.61 & 47.65 & 36.43 \\
BUTD-DETR \cite{jain2022bottom} & ECCV2022 & 84.20 & 66.30 & 46.60 & 35.10 & 52.20 & 39.80 \\
LAR \cite{bakr2022look} & NeurIPS2022 & - & - & - & - & 42.14 & 26.96 \\
EDA \cite{wu2023eda} & CVPR2023 & 85.76 & 68.57 & 49.13 & 37.64 & 54.59 & 42.26 \\
MA\textsuperscript{\tiny 2}TransVG \cite{xu2024multi} & CVPR2024 & 86.30 & 74.10 & \textbf{53.80} & 41.40 & 57.90 & 45.70 \\
ConcreteNet \cite{unal2024four} & ECCV2024 & 86.40 & \textbf{82.10} & 42.40 & 38.40 & 50.60 & 46.50 \\

\midrule

\multicolumn{8}{l}{\textit{3D Multimodal Large Language Models}} \\
VLM-Grounder \cite{xu2024vlm} & arXiv2024 & 66.00 & 29.80 & 48.30 & 33.50 & 51.60 & 32.80 \\
SeeGround \cite{li2025seeground} & CVPR2025 & 75.70 & 68.90 & 34.00 & 30.00 & 44.10 & 39.40 \\
View-on-Graph \cite{liu2026view} & arXiv2025 & 78.60 & 71.50 & 34.00 & 30.40 & 44.80 & 40.30 \\
ChatScene \cite{huang2024chat} & NeurIPS2024 & - & - & - & - & 55.50 & 50.20 \\
LLaVA-3D \cite{zhu2024llava} & arXiv2024 & - & - & - & - & 50.10 & 42.70 \\
Inst3D-LLM \cite{yu2025inst3d} & CVPR2025 & \textbf{88.60} & 81.50 & 48.70 & \textbf{43.20} & 57.80 & 51.60 \\
Robin3D \cite{kang2025robin3d} & ICCV2025 & - & - & - & - & 60.80 & 55.10 \\
Video-3D LLM \cite{zheng2025video} & CVPR2025 & - & - & - & - & 58.10 & 51.70 \\
GPT4Scene \cite{qi2025gpt4scene} & arXiv2025 & - & - & - & - & 62.60 & \textbf{57.00} \\
3DGraphLLM \cite{zemskova20253dgraphllm} & ICCV2025 & - & - & - & - & 62.40 & 56.60 \\
Vid-LLM \cite{chen2025vid} & arXiv2025 & - & - & - & - & \textbf{63.20} & 56.40 \\
Ross3D \cite{wang2025ross3d} & ICCV2025 & - & - & - & - & 61.10 & 54.40 \\
VG-LLM \cite{zheng2026learning} & ICLR2026 & - & - & - & - & 57.60 & 50.90 \\
3D-RFT \cite{linghu20263d} & arXiv2026 & - & - & - & - & 54.60 & 48.10 \\

\bottomrule
\end{tabular}}
\end{table*}

\begin{table*}[t]
\centering
\caption{Performance comparison of 3D dense captioning approaches on ScanRefer benchmark. CIDEr, BLEU-4, METEOR, and ROUGE, are used for evaluation.}
\label{tab:dense_captioning}
\scriptsize
\setlength{\tabcolsep}{5.2pt}
\renewcommand{\arraystretch}{1.08}
\resizebox{\textwidth}{!}{
\begin{tabular}{ll|rrrr|rrrr}
\toprule
\multirow{2}{*}{Method} & \multirow{2}{*}{Venue}
& \multicolumn{4}{c|}{IoU=0.25}
& \multicolumn{4}{c}{IoU=0.50} \\
\cmidrule(lr){3-6} \cmidrule(lr){7-10}
& & CIDEr & BLEU-4 & METEOR & ROUGE & CIDEr & BLEU-4 & METEOR & ROUGE \\
\midrule

\multicolumn{10}{l}{\textit{3D Dense Captioning Expert Models}} \\

SpaCap3D \cite{wang2022spatiality} & IJCAI2022
& - & - & - & -
& 42.76 & 25.38 & 22.84 & 45.66 \\

MORE \cite{jiao2022more} & ECCV2022
& 58.89 & 35.41 & 26.36 & 55.41
& 38.98 & 23.01 & 21.65 & 44.33 \\

X-Trans2Cap \cite{yuan2022x} & CVPR2022
& 58.81 & 34.17 & 25.81 & 54.10
& 41.52 & 23.83 & 21.90 & 44.97 \\


Vote2Cap-DETR \cite{chen2023end} & CVPR2023
& 71.45 & 39.34 & 28.25 & 59.33
& 61.81 & 34.46 & 26.22 & 54.40 \\

Vote2Cap-DETR++ \cite{chen2024vote2cap} & IEEE TPAMI2024
& 76.36 & 41.37 & 28.70 & 60.00
& 67.58 & 37.05 & 26.89 & 55.64 \\

BiCA \cite{kim2024bi} & ECCV2024
& 78.42 & \textbf{41.46} & 28.82 & 60.02
& 68.46 & \textbf{38.23} & 27.56 & \textbf{58.56} \\

\midrule

\multicolumn{10}{l}{\textit{Joint 3D Grounding and Captioning}} \\
D3Net \cite{chen2022d} & ECCV2022
& - & - & - & -
& 61.50 & 35.05 & 25.48 & 53.31 \\

3DJCG \cite{cai20223djcg} & CVPR2022
& 60.86 & 39.67 & 27.45 & 59.02
& 47.68 & 31.53 & 24.28 & 51.08 \\

PQ3D \cite{zhu2024unifying} & ECCV2024
& \textbf{86.40} & 38.80 & \textbf{30.70} & \textbf{61.00}
& \textbf{79.80} & 35.50 & \textbf{28.80} & 57.30 \\

\bottomrule
\end{tabular}}
\end{table*}

\begin{table*}[t]
\centering
\caption{Performance comparison on 3D question answering benchmarks. EM and EM-R are reported on SQA3D. CIDEr, BLEU-4, METEOR, ROUGE, and EM are reported on ScanQA.}
\label{tab:3d_qa}
\tiny
\setlength{\tabcolsep}{2.4pt}
\renewcommand{\arraystretch}{1.08}
\resizebox{\textwidth}{!}{
\begin{tabular}{llcc|cc|rrrrr}
\toprule
Method & Venue & Point & Vision
& \multicolumn{2}{c|}{SQA3D}
& \multicolumn{5}{c}{ScanQA} \\
& & Encoder & Encoder
& EM & EM-R
& CIDEr & BLEU-4 & METEOR & ROUGE & EM \\
\midrule

\multicolumn{11}{l}{\textit{3D Question Answering Expert Models}} \\
SQA3D \cite{ma2022sqa3d} & ICLR2023 & \checkmark & --
& 47.2 & --
& -- & -- & -- & -- & -- \\

ScanQA \cite{azuma2022scanqa} & CVPR2022 & \checkmark & --
& -- & --
& 64.9 & 10.1 & 13.1 & 33.3 & 21.1 \\

SIG3D \cite{man2024situational} & CVPR2024 & \checkmark & --
& 52.6 & --
& 68.8 & 12.4 & 13.4 & 35.9 & -- \\

3DGraphQA \cite{wu20243d} & arXiv2024 & \checkmark & \checkmark
& 49.2 & --
& -- & -- & -- & -- & -- \\

DSPNet \cite{luo2025dspnet} & CVPR2025 & \checkmark & \checkmark
& 50.4 & --
& -- & -- & -- & -- & -- \\


DTC \cite{huang2025zero} & CVPR2025 & -- & \checkmark
& -- & --
& -- & -- & -- & -- & 27.8 \\

SeGPruner \cite{li2026segpruner} & arXiv2026 & -- & \checkmark
& -- & --
& -- & -- & -- & -- & 28.0 \\

\midrule

\multicolumn{11}{l}{\textit{3D Multimodal Large Language Model}} \\
3D-LLM \cite{hong20233d} & NeurIPS2023 & \checkmark & \checkmark
& -- & --
& 69.4 & 12.0 & 14.5 & 35.7 & 20.5 \\

LL3DA \cite{chen2024ll3da} & CVPR2024 & \checkmark & --
& -- & --
& 76.8 & 13.5 & 15.9 & 37.3 & -- \\


Scene-LLM \cite{fu2025scene}  & arXiv2025 & \checkmark & \checkmark
& 54.2 & --
& 80.0 & 12.0 & 16.6 & 40.0 & 27.2 \\

ChatScene \cite{huang2024chat} & NeurIPS2024 & \checkmark & \checkmark
& 54.6 & 57.5
& 87.7 & 14.3 & 18.0 & 41.6 & 21.6 \\

ChatScene++ \cite{huang2026chat} & TPAMI2026 & \checkmark & \checkmark
& 55.8 & 58.5
& 93.5 & 18.4 & -- & -- & -- \\

Robin3D \cite{kang2025robin3d} & ICCV2025 & \checkmark & --
& 56.9 & 59.8
& -- & -- & 19.2 & 44.0 & -- \\

Inst3D-LLM \cite{yu2025inst3d} & CVPR2025 & \checkmark & \checkmark
& -- & --
& 88.6 & 14.9 & 18.4 & 42.6 & 24.6 \\

LLaVA-3D \cite{zhu2024llava}  & ICCV2025 & -- & \checkmark
& 60.1 & --
& 103.1 & 16.4 & 20.8 & 49.6 & 30.6 \\

S$^2$-MLLM \cite{xu2025s} & CVPR2026 & -- & \checkmark
& 55.9 & 58.7
& 91.8 & 14.8 & 18.4 & 45.0 & -- \\

3DGraphLLM \cite{zemskova20253dgraphllm} & ICCV2025 & \checkmark & --
& 55.9 & --
& 88.8 & 15.9 & -- & -- & -- \\

SplatTalk \cite{thai2025splattalk} & ICCV2025 & -- & \checkmark
& 54.5 & --
& 101.9 & 17.1 & \textbf{23.2} & \textbf{52.4} & -- \\

PoseAlign \cite{liu2026direction} & arXiv2026 & \checkmark & --
& 54.2 & 56.7
& 99.8 & 17.3 & 19.7 & 46.5 & -- \\

GPT4Scene \cite{qi2025gpt4scene} & ICLR2026 & -- & \checkmark
& -- & --
& 96.3 & 15.5 & 18.9 & 46.5 & 28.2 \\

Video-3D LLM \cite{zheng2025video} & CVPR2025 & -- & \checkmark
& 58.6 & --
& 102.1 & 16.4 & 20.0 & 49.3 & 30.1 \\

Vid-LLM \cite{chen2025vid} & arXiv2025 & -- & \checkmark
& 60.1 & 63.0
& \textbf{107.8} & 16.4 & 20.9 & 50.3 & \textbf{30.8} \\

Ross3D \cite{wang2025ross3d} & ICCV2025 & -- & \checkmark
& \textbf{63.0} & \textbf{65.7}
& 107.0 & 17.9 & 20.9 & 50.7 & \textbf{30.8} \\

SAGE3D \cite{paul2026point} & CVPR2026 & \checkmark & --
& -- & --
& -- & 17.5 & 19.6 & 44.5 & -- \\

ENEL \cite{tang2025exploring} & arXiv2025 & -- & --
& -- & --
& 52.7 & -- & 13.5 & 31.2 & -- \\
\bottomrule
\end{tabular}}
\end{table*}

\begin{table*}[t]
\centering
\caption{Performance comparison of open-vocabulary 3D recognition approaches on ScanNet benchmark. The results are reported under diverse base/novel splits and annotation-free or zero-shot protocols.}
\label{tab:ov_3d_semseg}
\tiny
\setlength{\tabcolsep}{2.6pt}
\renewcommand{\arraystretch}{1.08}
\resizebox{\textwidth}{!}{
\begin{tabular}{ll|rrr|rrr|rrr|rrrr}
\toprule
Method & Venue
& \multicolumn{3}{c|}{B15/N4}
& \multicolumn{3}{c|}{B12/N7}
& \multicolumn{3}{c|}{B10/N9}
& \multicolumn{4}{c}{Annotation-free / Zero-shot} \\
&
& hIoU & mIoU$_B$ & mIoU$_N$
& hIoU & mIoU$_B$ & mIoU$_N$
& hIoU & mIoU$_B$ & mIoU$_N$
& mIoU & mAcc & f-mIoU & f-mAcc \\
\midrule

Lowis3D \cite{ding2024lowis3d} & TPAMI2024
& 65.3 & 68.3 & 62.4
& 55.3 & 69.5 & 45.9
& 53.1 & 76.2 & 40.8
& - & - & - & - \\

RegionPLC \cite{yang2024regionplc} & CVPR2024
& 69.4 & 68.2 & 70.7
& 68.2 & 69.9 & 66.6
& 64.3 & 76.3 & 55.6
& - & - & 43.8 & 65.6 \\

OV3D \cite{jiang2024open} & CVPR2024
& 72.4 & 70.2 & 74.7
& 68.5 & 74.1 & 63.7
& 64.8 & 77.6 & 55.6
& 57.3 & 72.9 & 64.0 & 76.3 \\

UniM-OV3D \cite{he2024unim} & IJCAI2024
& \textbf{75.8} & \textbf{75.3} & \textbf{76.6}
& \textbf{74.7} & \textbf{75.2} & \textbf{74.1}
& \textbf{69.9} & \textbf{83.5} & \textbf{57.3}
& - & - & - & - \\

CLIP2Scene \cite{chen2023clip2scene} & CVPR2023
& - & - & -
& - & - & -
& - & - & -
& 25.08 & - & - & - \\

CLIP-FO3D & ICCVW2023
& - & - & -
& - & - & -
& - & - & -
& 30.2 & 49.1 & - & - \\

OpenScene \cite{peng2023openscene} & CVPR2023
& - & - & -
& - & - & -
& - & - & -
& 54.2 & 66.6 & - & - \\

OV3D + OpenScene-3D \cite{jiang2024open} & CVPR2024
& - & - & -
& - & - & -
& - & - & -
& \textbf{59.6} & \textbf{74.5} & \textbf{67.6} & 79.1 \\


MPEC \cite{wang2025masked} & CVPR2025
& - & - & -
& - & - & -
& - & - & -
& - & - & 66.0 & \textbf{81.3} \\
\bottomrule
\end{tabular}}
\end{table*}

\begin{table*}[t]
\centering
\caption{Performance comparison of VLA models on LIBERO, RLBench, and COLOSSEUM benchmarks. Success rate is reported on four LIBERO suites, including Spatial, Object, Goal, and Long. Average success rate (Avg. SR) and average rank (Avg. Rank) are reported on RLBench and COLOSSEUM benchmarks.}
\label{tab:embodied}
\tiny
\setlength{\tabcolsep}{2.6pt}
\renewcommand{\arraystretch}{1.08}
\resizebox{\textwidth}{!}{
\begin{tabular}{ll|rrrrr|rr|rr}
\toprule
Method & Venue
& \multicolumn{5}{c|}{LIBERO}
& \multicolumn{2}{c|}{RLBench}
& \multicolumn{2}{c}{COLOSSEUM} \\
&
& Spatial & Object & Goal & Long & Avg.
& Avg. SR & Avg. Rank
& Avg. SR & Avg. Rank \\
\midrule

SpatialVLA \cite{qu2025spatialvla} & RSS2025
& 88.2 & 89.9 & 78.6 & 55.5 & 78.1
& - & - & - & - \\

3D-CAVLA \cite{bhat20253d} & CVPRW2025
& 98.2 & \textbf{99.8} & 98.2 & 96.1 & 98.1
& - & - & - & - \\

GeoVLA \cite{sun2025geovla} & arXiv2025
& 98.4 & 99.0 & 96.6 & \textbf{96.6} & 97.7
& - & - & - & - \\

SpatialForcing \cite{li2025spatial} & ICLR2026
& \textbf{99.4} & 99.6 & 98.8 & 96.0 & 98.5
& - & - & - & - \\

GST-VLA \cite{sarowar2026gst} & arXiv2026
& 98.2 & 97.4 & 97.1 & 92.6 & 96.4
& - & - & - & - \\

ANY3D-VLA \cite{fan2026any3d} & arXiv2026
& 81.2 & 85.8 & 95.8 & 86.2 & 87.3
& - & - & - & - \\

ABot-M0 \cite{yang2026abot} & arXiv2026
& 98.8 & \textbf{99.8} & \textbf{99.0} & \textbf{96.6} & \textbf{98.6}
& - & - & - & - \\



RVT \cite{goyal2023rvt} & CoRL2023
& - & - & - & - & -
& 62.9 & 1.1 & - & - \\

Act3D \cite{gervet2023act3d} & CoRL2023
& - & - & - & - & -
& 65.0 & - & - & - \\

PolarNet \cite{chen2023polarnet} & CoRL2023
& - & - & - & - & -
& 46.4 & - & - & - \\

3D-DiffuserActor \cite{ke20243d} & CoRL2024
& - & - & - & - & -
& 81.3 & - & - & - \\

3DS-VLA \cite{li20253ds} & CoRL2025
& - & - & - & - & -
& 66.0 & - & - & - \\

Evo-0 \cite{lin2025evo} & arXiv2025
& - & - & - & - & -
& 56.0 & - & - & - \\

BridgeVLA \cite{li2026bridgevla} & NeurIPS2025
& - & - & - & - & -
& 88.2 & 1.9 & 64.0 & \textbf{1.07} \\

ActiveVLA  \cite{liu2026activevla} & CVPR2026
& - & - & - & - & -
& \textbf{91.8} & \textbf{1.02} & \textbf{65.9} & \textbf{1.07} \\



\bottomrule
\end{tabular}}
\end{table*}

\section{Performance Evaluation}
\label{section5}

\subsection{Datasets}

\textbf{KITTI} \cite{geiger2012we} is one of the most popular benchmark datasets for 3D object detection and orientation estimation in autonomous driving, which comprises more than 200,000 annotated point cloud scenarios consisting of cars and pedestrians. This dataset includes not only point cloud but also stereo and optical flow data, which is collected using several pieces of equipment, including four video cameras, a laser scanner, and a localization system. Specifically, 7,481 samples with annotations in the camera field of vision are used for training and the other 7,518 samples are remained for testing. According to the occlusion level, visibility, and bounding box size, the samples are further divided into three difficulty levels, including easy, moderate, and hard.

\textbf{nuScenes} \cite{caesar2020nuscenes} is another multi-modal autonomous driving dataset provided by the full sensor suite, including cameras, radars, and LiDARs. Specifically, the nuScenes dataset contains 1,000 scenes. Each scene has approximately 40,000 frames 
, where 28,130/6,019/6,008 frames are regarded as training/validation/ testing samples. For 3D object detection, after merging similar classes and removing rare classes, the key samples are annotated with ground truth labels of 10 foreground object classes at a frequency of 2Hz. For 3D semantic segmentation, every point in the keyframe is annotated using 6 more background classes in addition to the 10 foreground object classes.

\textbf{SemanticKITTI} \cite{behley2019semantickitti} is a large-scale outdoor-scene dataset for 3D semantic segmentation, which annotates all data of KITTI Vision Odometry Benchmark \cite{geiger2012we} and provides dense point-wise annotations for the complete 360 field-of-view of the employed automotive LiDAR. For consistency with the KITTI Vision Odometry Benchmark, the SemanticKITTI dataset employs the same division for our training and test set. It is worth noting that SemanticKITTI provides as many as 28 classes, but the official evaluation only uses a set of 19 valid classes due to ignoring classes with only a few points and merging classes with different mobility states.

\textbf{Waymo} \cite{sun2020scalability} is a dataset for autonomous driving perception. It contains 1,150 driving scenes, each lasting 20 seconds, collected from urban and suburban areas in San Francisco, Phoenix, and Mountain View. The dataset provides well-synchronized and calibrated data from five LiDAR sensors and five high-resolution cameras. It includes around 12 million 3D LiDAR bounding boxes and around 12 million 2D camera bounding boxes, with consistent object identities across frames. The annotations cover vehicles, pedestrians, cyclists, and traffic signs, supporting 2D and 3D object detection as well as multi-object tracking. 

\textbf{ScanNet} \cite{dai2017scannet} is a large-scale RGB-D dataset for 3D indoor scene understanding and reconstruction. It contains 1,513 scans covering 707 unique indoor spaces, including offices, apartments, bathrooms, classrooms, and libraries. Each scan consists of RGB-D frames with corresponding camera poses. In addition, ScanNet provides instance-level semantic annotations for scanned objects, covering common indoor categories such as chairs, tables, sofas, and beds.

\textbf{ScanRefer} \cite{chen2020scanrefer} is a large-scale dataset for 3D visual grounding and dense captioning. The dataset contains 51,583 human-written free-form descriptions of 11,046 objects based on the ScanNet scenes \cite{dai2017scannet}. The object descriptions cover more than 250 common indoor objects, such as sofas, chairs, table lamps, etc., and include information about the object's appearance and spatial relationship with other objects. ScanRefer dataset follows the ScanNet benchmark to split the train/val/test sets with 36,665, 9,508, and 5,410 samples, respectively.

\textbf{ScanQA} \cite{azuma2022scanqa} is a large-scale dataset for 3D object-grounded question answering on point cloud. It consists of 41,363 questions and 58,191 answers, including 32,337 unique questions and 16,999 unique answers. Various types of questions in the dataset are collected through auto-generation and editing by humans. Alongside question-answer pairs, the dataset also includes 3D object localization annotations for 800 indoor 3D scenes of the ScanNet dataset. There are two test sets with and without object annotations in ScanQA dataset.

\textbf{SQA3D} \cite{ma2022sqa3d} is the largest dataset of grounded 3D scene understanding with the human-annotated question-answering pairs. It comprises 20.4k descriptions of 6.8k unique situations collected from 650 ScanNet scenes and 33.4k questions about these situations. These questions examine a wide spectrum of reasoning capabilities for an intelligent agent, ranging from spatial relation comprehension to commonsense understanding, navigation, and multi-hop reasoning. The dataset is divided into three splits, with a total of 518/65/67 scenes and 11,723/1,550/1,652 targets across the train/val/test sets, respectively.

\textbf{LIBERO} \cite{liu2023libero} is a benchmark for lifelong robot learning in manipulation scenarios. This benchmark contains 130 language-conditioned robot manipulation scenarios generated through a procedural pipeline, covering diverse objects, spatial arrangements, backgrounds, and manipulation goals. The benchmark is divided into four suites. LIBERO-Spatial, LIBERO-Object, and LIBERO-Goal encompass 10 scenarios for evaluating spatial, object-level, and goal-level knowledge transfer, respectively. LIBERO-100 contains 100 scenarios with more diverse and entangled knowledge transfer (10 scenarios for long-horizon manipulation). 

\textbf{RLBench} \cite{james2020rlbench} is a large-scale benchmark and learning environment for robot manipulation. It contains 100 completely unique hand-designed scenarios with different levels of difficulty, ranging from simple target reaching and door opening to long-horizon multi-stage manipulations such as opening an oven and placing a tray inside it. Each scenario provides multiple linguistic descriptions and supports an infinite number of demonstrations generated by motion planners based on predefined waypoints. The observations include RGB images, depth maps, and segmentation masks from an over-the-shoulder stereo camera and an eye-in-hand monocular camera, together with proprioceptive information such as joint angles, velocities, torques, and gripper poses.

\textbf{COLOSSEUM} \cite{pumacay2024colosseum} is a simulation benchmark for evaluating generalization in robotic manipulation. The benchmark systematically evaluates robot policies under 14 types of environmental perturbations, including changes in object color, texture, and size, table appearance, background texture, lighting condition, distractor objects, camera pose, object friction, and object mass. In total, COLOSSEUM provides more than 20,000 unique perturbed instances and supports both individual and combined perturbation settings. 

\subsection{Evaluation Metrics}
\textbf{Evaluation Metrics on 3D Semantic Segmentation.} 
For semantic segmentation of 3D point cloud, mean Intersection-over-Union (mIoU) is the most frequently used performance criteria, which is defined as ${\frac{1}{C}\sum_{c=1}^{C}{\frac{{\rm TP}_c/}{{\rm TP}_c+{\rm FP}_c+{\rm FN}_c}}}$, where ${\rm TP}_c$, ${\rm FP}_c$, and ${\rm FN}_c$ correspond to the number of true positive, false positive, and false negative predictions for class $c$, and $C$ is the number of classes. Besides, frequency-weighted IOU (FwIOU) is also used in 3D semantic segmentation, which is similar to mIoU except that each IoU is weighted by the point-level frequency of its class.

\textbf{Evaluation Metrics on 3D Object Detection.} Average Precision (AP) is the main evaluation metric used in 3D object detection, which is calculated as the area under the precision-recall curve \cite{padilla2020survey}. Based on the original definition of AP, some dataset-specific evaluation metrics are developed for different datasets. For the KITTI dataset, Average Orientation Similarity (AOS) is introduced to assess the accuracy of orientation estimation \cite{geiger2012we}. For the nuScenes dataset, nuScenes detection score (NDS) is the major evaluation metric.

\textbf{Evaluation Metrics on 3D Visual Grounding.}
In 3D VG, the evaluation metric is the Acc$@m$IoU, which means the fraction of descriptions whose predicted box overlaps the ground truth with IoU $> m$, where $m\in\{0.25, 0.5\}$ \cite{chen2020scanrefer}.The accuracy is reported in unique and multiple categories, where a target object is classified as unique if it is the only object of its class in the scene. Otherwise, it is classified as multiple. 

\textbf{Evaluation Metrics on 3D Dense Captioning.}
For  3D dense captioning, the most commonly used evaluation metric is $m@k$IoU \cite{chen2021scan2cap}, which enables a holistic assessment of both the quality of the generated descriptions and the detected bounding boxes. $m@k$IoU is defined as $m@k$IoU$={\frac{1}{N}}\sum_{i=0}^{N}{m_iu_i}$, where $u_i\in\{0,1\}$ is set to 1 if the IoU score for the $i^{th}$ box is greater than $k$, otherwise 0. In the metric, $m$ represents the captioning metrics CiDEr \cite{vedantam2015cider}, BLEU-4 \cite{papineni2002bleu}, METEOR \cite{banerjee2005meteor}, and ROUGE  \cite{lin2004rouge}, abbreviated as C, B-4, M, R, respectively. $N$ is the number of ground truth or detected object bounding boxes.

\textbf{Evaluation Metrics on 3D Question Answering.}
To evaluate the performance of 3D QA, some frequently used evaluation metrics are EM@1 and EM@10, where EM@$K$ represents the percentage of predictions in which the top $K$ predicted answers exactly match any one of the ground-truth answers \cite{rajpurkar2016squad}. Additionally, some sentence evaluation metrics commonly used for 3D dense captioning, including BLEU \cite{papineni2002bleu}, ROUGE, METEOR, CIDEr, and SPICE \cite{anderson2016spice} are added to analyze the robust answer matching, as some questions have multiple possible answer expressions.

\textbf{Evaluation Metrics on Embodied Understanding.}
For embodied understanding, success rate (SR) is the primary evaluation metric, denoting the percentage of episodes in which an agent successfully completes the target task. On LIBERO  \cite{liu2023libero}, SR is reported on four suites, including Spatial, Object, Goal, and Long, which respectively assess spatial reasoning, object-level recognition, goal-conditioned manipulation, and long-horizon execution. The average SR over these suites is further used to measure the overall performance. On RLBench \cite{james2020rlbench} and COLOSSEUM \cite{pumacay2024colosseum}, average SR is adopted to evaluate the completion ability across diverse manipulation tasks and perturbation conditions. In addition, average rank is used to measure the relative standing of each method among all compared approaches across tasks or benchmarks, where a lower rank indicates better overall performance.

\subsection{Performance Evaluation of Image-assisted 3D Scene Understanding}
To systematically compare 3D detection approaches, Table~\ref{tab:kitti_detection} and Table~\ref{tab:nuscenes_detection} summarize representative works on the KITTI \cite{geiger2012we} and nuScenes \cite{caesar2020nuscenes} benchmarks. KITTI evaluates ``car”, ``pedestrian”, and ``cyclist” detection, whereas nuScenes further reflects performance under more categories and complex scenes. In summary, evaluation results demonstrate that detection performance is strongly correlated with the capacity of the strategy to fuse complementary 2D and 3D information across modalities. The main observations are presented as follows:

\begin{itemize}
    \item Hierarchical cross-modal fusion achieves consistently strong performance. On the KITTI \cite{geiger2012we} benchmark, LoGoNet \cite{li2023logonet} achieves the best or highly competitive mAP across the ``car”, ``pedestrian”, and ``cyclist” categories, with particularly notable gains on the ``cyclist”class. Compared with simple camera-LiDAR projection or global feature concatenation, hierarchical global-to-local fusion simultaneously models scene-level context and object-level local details, thereby improving detection robustness under challenging scenarios involving occlusion and small objects.
    
    \item Attention mechanism provides more stable cross-modal interaction in complex scenes. Compared with camera-LiDAR projection fusion, Transformer 3D detection architectures generally achieve stronger mAP and NDS on nuScenes \cite{caesar2020nuscenes}. In nuScenes, 3D scenes contain more object categories and larger scale variations. These results indicate that adaptive fusion is essential for cross-modal 3D detection, because it can dynamically emphasize informative visual or geometric cues while suppressing noisy or less relevant features. In addition, Transformer architectures also enhance long-range dependency modeling across modalities.

    \item Temporal modeling further improves 3D detection and is a crucial direction for future research. Historical information on the location and orientation of moving objects can improve estimation of their current motion. In addition, temporal cues from adjacent frames can facilitate detection of distant or occluded objects by providing complementary information. However, directly aggregating 4D spatiotemporal data, as in LIFT \cite{zeng2022lift}, often yields suboptimal performance, primarily because dense 4D representations contain substantial redundancy, complicating alignment of objects across sensors and timestamps. Therefore, modeling the spatial and temporal domains separately, as in BEVFusion4D \cite{cai2023bevfusion4d}, represents a promising research direction.
\end{itemize}

Table~\ref{tab:semantic_segmentation} compares 3D semantic segmentation approaches across outdoor autonomous-driving benchmarks, including SemanticKITTI \cite{behley2019semantickitti}, nuScenes \cite{caesar2020nuscenes}, and Waymo \cite{sun2020scalability}, and indoor RGB-D benchmarks, ScanNet \cite{dai2017scannet}. The main observations are presented as follows:
\begin{itemize}
    \item For interactive fusion approaches, addressing the limited intersection on the FOV between sensors is important to improve the overall performance. This is because only 3D points falling into the intersected FOV are geometrically associated with 2D pixels, while only considering the intersected cross-modal data may discard a lot of useful information in those unmatched point-pixel pairs. Overcoming the inapplicable and overlooked LiDAR-camera fusion outside the sensor FOV intersection, like MSeg3D \cite{li2023mseg3d}, can bring huge performance improvements. Inspired by this observation, cross-modal feature completion and semantics-aware feature fusion emerge as promising directions for addressing incomplete sensor overlap.
    
    \item On outdoor autonomous-driving benchmarks, cross-modal knowledge distillation shows more consistent advantages in large-scale sparse 3D scenarios. Specifically, U2MKD obtains the highest test mIoU on SemanticKITTI and consistently leads on nuScenes across Test FwIoU, Val mIoU, and Test mIoU. This reflects the limitations of direct point-pixel fusion in outdoor scenes, where sparse LiDAR, occlusions, and limited field-of-view overlap can weaken explicit cross-modal alignment.  By distilling 2D semantic priors into the 3D encoder, robustness can be improved under incomplete view observations, thereby alleviating the performance bottleneck caused by imperfect sensor overlap.
\end{itemize}

\subsection{Performance Evaluation of 3D Vision-Language Scene Understanding}
To provide a comprehensive evaluation of 3D visual-language reasoning approaches, we summarize the quantitative results on 3D visual grounding, 3D dense captioning, and 3D question answering. Specifically, Table~\ref{tab:visual_grounding} compares localization accuracy on ScanRefer \cite{chen2020scanrefer}, Table~\ref{tab:dense_captioning} reports captioning quality on ScanRefer, and Table~\ref{tab:3d_qa} presents question answering performance on SQA3D \cite{ma2022sqa3d} and ScanQA \cite{azuma2022scanqa}. The main observations are presented as follows:

\begin{itemize}
    \item Expert models continue to demonstrate strong performance in dense object localization and scene description. As shown in Table~\ref{tab:visual_grounding}, for 3D visual grounding, approaches such as EDA \cite{wu2023eda}, MA\textsuperscript{\tiny 2}TransVG \cite{xu2024multi}, and ConcreteNet \cite{unal2024four} achieve high accuracy on ScanRefer under both Acc@0.25 and Acc@0.5 metrics. Similarly, as shown in Table~\ref{tab:dense_captioning}, for 3D dense captioning, BiCA \cite{kim2024bi} consistently outperforms other approaches on standard metrics including CIDEr, METEOR, and ROUGE-L. These observations highlight that specialized architectures can provide reliable baselines for closed-set 3D visual-language reasoning applications with well-defined evaluation protocols.
    
    \item 3D MLLMs demonstrate strong general semantic reasoning ability, but accurate spatial alignment remains a key limitation. As shown in Table~\ref{tab:visual_grounding}, several 3D MLLMs achieve better performance than expert models on the overall metrics for visual grounding. This suggests that the semantic understanding inherited from large language models can benefit 3D scene parsing. However, their advantages are less consistent under the ``unique” and ``multiple” settings, especially when multiple objects with similar semantics appear in the same scene. This observation indicates that current 3D MLLMs still face challenges in precise object localization within complex 3D environments.

    \item The advantage of 3D MLLMs is more pronounced in open ended 3D QA. As reported on SQA3D \cite{ma2022sqa3d} and ScanQA \cite{azuma2022scanqa}, recent 3D MLLMs generally achieve higher scores than 3D QA expert models. These results suggest that open ended 3D QA requires not only geometric perception, but also language understanding, contextual reasoning, and commonsense knowledge. Such requirements are naturally aligned with the strengths of MLLM architectures. Besides, vision enhanced approaches achieve the strongest results. For example, Ross3D \cite{wang2025ross3d} obtains the best SQA3D performance with 63.0 EM and 65.7 EM-R. Vid-LLM \cite{chen2025vid} achieves the highest CIDEr score of 107.8 and ties for the best EM score of 30.8 on ScanQA. These results suggest that incorporating richer video cues can provide more informative representations for open ended 3D QA.
\end{itemize}

To systematically compare open-vocabulary 3D recognition approaches, Table~\ref{tab:ov_3d_semseg} summarizes representative approaches on indoor ScanNet \cite{dai2017scannet} datasets. Based on ScanNet, researchers evaluate base/novel category generalization under different class splits. The annotation-free and zero-shot settings further assess whether models can recognize open-vocabulary categories without manual 3D annotations.
\begin{itemize}
    \item Strong open-vocabulary 3D recognition requires enhancing the semantic discriminability of 3D representations, rather than simply leveraging category priors from 2D foundation models. On the ScanNet \cite{dai2017scannet} base/novel splits, UniM-OV3D \cite{he2024unim} achieves the highest hIoU across B15/N4, B12/N7, and B10/N9, with scores of 75.8, 74.7, and 69.9, respectively. Performance gains primarily stem from the fact that UniM-OV3D aligns point clouds, images, depth, and text in a unified feature space and introduces hierarchical point-level semantic supervision. This enables 3D point features to better respond to unseen-category queries.

    \item In annotation-free and zero-shot scenarios, 3D recognition depends on grounding 2D semantic priors into geometrically consistent 3D representations. For example, OV3D + OpenScene-3D \cite{jiang2024open} obtains 59.6 mIoU, 74.5 mAcc, 67.6 f-mIoU, and 79.1 f-mAcc on ScanNet \cite{dai2017scannet}. MPEC \cite{wang2025masked} further improves f-mAcc to 81.3. These results indicate that 2D foundation models provide rich open-vocabulary semantics, but their predictions must be constrained by 2D-3D correspondences and spatially coherent regions. Such 3D constraints reduce errors caused by viewpoint variation and noisy projection, enabling semantic priors to be more reliably assigned to 3D points and regions without manual 3D annotations.
\end{itemize}

\subsection{Performance Evaluation of 3D Embodied Understanding}

Table~\ref{tab:embodied} presents representative 3D VLA frameworks evaluated on LIBERO \cite{liu2023libero}, RLBench \cite{james2020rlbench}, and COLOSSEUM \cite{pumacay2024colosseum}. LIBERO focuses on spatial reasoning, object manipulation, goal-directed actions, and long-horizon perception. RLBench \cite{james2020rlbench} and COLOSSEUM \cite{pumacay2024colosseum} evaluate performance under multi-view and partially occluded conditions. The main observations are presented as follows:
\begin{itemize}
    \item On LIBERO \cite{liu2023libero} benchmark, 3D spatial modeling substantially improves the performance of embodied manipulation. In particular, SpatialVLA \cite{qu2025spatialvla} achieves an average success rate of 78.1\%, with only 55.5\% on the Long split, while recent 3D-aware methods such as SpatialForcing \cite{li2025spatial}, ABot-M0 \cite{yang2026abot}, and 3D-MIX \cite{yu20263d} reach around 98\% average success rates. This suggests that explicit 3D spatial representations and action-manifold modeling enhance the model’s ability to reason about object locations, goal states, and action constraints. However, the Long split remains more challenging than Spatial, Object, and Goal, indicating that long-horizon manipulation is still limited by long-range action-state and accumulated execution errors.

    \item The comparative results highlight the importance of active perception and action-relevant viewpoint selection in complex manipulation scenarios. In particular, ActiveVLA \cite{liu2026activevla} achieves 91.8 Avg. SR on RLBench \cite{james2020rlbench} and 65.9 Avg. SR on COLOSSEUM \cite{pumacay2024colosseum}, outperforming other VLA frameworks on both benchmarks. This suggests that fixed-view 3D perception alone is insufficient for manipulation under occlusion, viewpoint changes, and fine-grained interaction requirements. By localizing critical 3D regions, actively selecting informative viewpoints, and applying 3D zoom-in, ActiveVLA better aligns perceptual inputs with action prediction, improving manipulation generalization. In the future, embodied VLA models should integrate active perception and action feedback into a unified closed-loop decision-making framework.
\end{itemize}

\section{Future Directions}
Despite remarkable progress in 3D understanding and generation, existing methods remain limited in their ability to scale, generalize, interact, and support embodied decision making. We outline several future directions that may shape the next generation of 3D models, ranging from action-oriented perception to scalable representations and efficient systems.


\textbf{Persistent 3D memory.}
Another crucial direction is to move from short-term scene perception to persistent and updatable 3D memory. Real-world agents observe environments sequentially under partial views, occlusion, motion, and scene changes, making one-shot reconstruction insufficient. Future models should maintain compact spatial memories that integrate multi-view observations over time, distinguish static scene structure from dynamic objects, and update uncertainty as new evidence arrives. Such memories should support not only reconstruction, but also semantic querying, localization, navigation, and interaction. Persistent 3D memory can therefore provide a long-horizon scene representation that connects generation, perception, reasoning, and action.

\textbf{3D-aware physical world models.}
A further opportunity lies in building hybrid world models that combine the generative capacity of video or latent models with explicit 3D physical representations. While video world models can capture rich visual dynamics, their internal states are often implicit and difficult to query, edit, or constrain geometrically. In contrast, explicit 3D representations provide structured spatial states for reasoning about geometry, motion, contact, and physical interaction, but may lack the flexibility and visual realism of large generative models. Future work should bridge these paradigms by grounding generative world models in persistent 3D scene states, such as object-centric maps, scene graphs, neural fields, or Gaussian primitives, and updating them through action-conditioned dynamics. Such 3D-aware physical world models would support not only visual prediction, but also spatial reasoning, counterfactual simulation, planning, and embodied interaction.

\textbf{Scalable scene representations.}
A key challenge for future 3D modeling is to design representations that scale from individual objects to large, complex scenes while preserving fine geometric details, semantic structure, and interaction-relevant properties. Existing representations such as meshes, neural fields, pointmaps, and Gaussian primitives each offer different trade-offs in fidelity, efficiency, editability, and physical compatibility, but none fully satisfies the needs of large-scale reconstruction, generation, and interaction. Future work may move toward hierarchical and compositional representations that combine global scene layouts with object-level and part-level structures, while integrating explicit geometry with compact neural or generative latent features. Such scalable representations would enable efficient storage, rendering, editing, reasoning, and simulation across diverse real-world environments.

\textbf{Benchmarks and evaluation protocols.}
A critical but underexplored direction is to rethink how multi-modal 3D intelligence should be evaluated. Existing benchmarks are largely inherited from isolated tasks, such as 3D detection, segmentation, visual grounding, captioning, question answering, or scene generation. Although these protocols are useful for measuring task-specific progress, they often fail to reveal whether a model has acquired a coherent understanding of the physical scene. For example, a model may achieve high grounding accuracy by exploiting category or language priors, generate visually plausible scenes with incorrect geometry, or produce fluent answers without maintaining spatial consistency. This mismatch suggests that future benchmarks should move from task-level evaluation to capability-level evaluation. Rather than only asking whether a model detects an object or answers a question correctly, benchmarks should test whether it can maintain consistent geometry across views, distinguish observed regions from inferred or hallucinated content, reason about occlusion and spatial relations, preserve object identity over time, and support downstream interaction or planning. 

Future evaluation protocols should therefore include diagnostic tests for geometry faithfulness, language grounding, open-vocabulary generalization, temporal consistency, physical plausibility, and action-level utility. In particular, generation models should be evaluated not only by visual realism, but also by 3D consistency, editability, collision validity, and compatibility with simulation or embodied tasks. Similarly, embodied 3D models should be assessed by whether their scene representations improve navigation, manipulation, and long-horizon decision making, rather than by perception accuracy alone. Such capability-oriented benchmarks would provide a more faithful measurement of progress toward general 3D world intelligence.

\textbf{System efficiency.}
Improving system efficiency is essential for deploying 3D models in real-world applications such as robotics, AR/VR, and autonomous systems. Current methods often require high memory, dense computation, or expensive multi-view processing, which limits their use in streaming and resource-constrained settings. Future work should focus on lightweight architectures, compact scene representations, efficient cross-view fusion, and incremental updates that avoid redundant computation. Beyond faster inference, efficient 3D systems should support real-time reconstruction, rendering, querying, and interaction under limited hardware budgets.

\section{Conclusion}

This survey presents a comprehensive review of recent advances in multi-modal 3D intelligence, covering both scene understanding and generation. We first introduced the problem definitions, modality settings, and key challenges in 3D+2D and 3D+language scenarios. We then organized existing methods into a structured taxonomy according to their input modalities, target tasks, and technical designs, including 3D object detection, semantic segmentation, vision-language reasoning, open-vocabulary recognition, 3D multimodal large language models, text-driven scene generation, and embodied 3D understanding. In addition, we summarized representative datasets and benchmarks, and provided comparative results and analysis to highlight the strengths and limitations of different approaches. We hope this survey can serve as a useful reference for both newcomers and experienced researchers, offering a systematic understanding of the field and inspiring further progress.

\ifCLASSOPTIONcaptionsoff
  \newpage
\fi

\balance
\bibliographystyle{IEEEtran}
\bibliography{mybib}

\end{document}